%% file: arxiv.tex
\documentclass{article}
\usepackage[margin=1in]{geometry}
\usepackage[utf8]{inputenc}
\usepackage[numbers]{natbib}
\usepackage{authblk}

\include{preamble}
\include{defs}

\newif\ifpnas
\pnasfalse

\title{Conformal Prediction Under Feedback Covariate Shift\\for Biomolecular Design}
\author[a]{Clara Fannjiang}
\author[a,b]{Stephen Bates}
\author[a]{Anastasios N. Angelopoulos}
\author[a,c]{Jennifer Listgarten}
\author[a,b]{Michael I. Jordan}

\affil[a]{Department of Electrical Engineering and Computer Sciences, University of California, Berkeley, USA}
\affil[b]{Department of Statistics, University of California, Berkeley, USA}
\affil[c]{Center for Computational Biology, University of California, Berkeley, USA}

\date{\vspace{-3ex}}
\begin{document}

\maketitle

\begin{abstract}
\input{abstract}
\end{abstract}

\input{main}

\bibliographystyle{plainnat}
\bibliography{references}

\clearpage

\setcounter{figure}{0}
\setcounter{section}{0}
\renewcommand{\thefigure}{A\arabic{figure}}
\renewcommand{\thesection}{A\arabic{section}}
\renewcommand{\theHfigure}{appendix.\thefigure}
\renewcommand{\theHsection}{appendix.\thesection}

\input{appendix}

\end{document}

%% file: preamble.tex
\usepackage{amsmath,amsthm,amsfonts,amssymb}
\usepackage{bm, bbm}
\usepackage{xcolor, graphicx,xparse}
\usepackage{subcaption}
\usepackage{algpseudocode, algorithm}  % or above line
\usepackage{hyperref}
\usepackage{mathtools}
\usepackage{ifthen}
\mathtoolsset{showonlyrefs=true}

\usepackage{letltxmacro}
\LetLtxMacro{\originaleqref}{\eqref}
\renewcommand{\eqref}{Eq.~\originaleqref}

\newtheorem{theorem}{Theorem}
\newtheorem{corollary}{Corollary}
\newtheorem{lemma}{Lemma}

\newtheorem{definition}{Definition}

\newtheorem{proposition}{Proposition}

\newcommand{\reals}{\mathbb{R}}

\newcommand{\Prob}{\mathbb{P}}

\DeclarePairedDelimiterX{\infdivx}[2]{(}{)}{
  #1\;\delimsize|\delimsize|\;#2
}
\DeclareMathOperator*{\argmin}{arg\,min}

%% file: defs.tex
\newcommand{\augloodata}[1]{Z^y_{-#1}}

\newcommand{\augloomu}[1]{\mu^y_{-#1}}
\newcommand{\quantile}[1]{\textsc{Quantile}_{#1}}

%% file: abstract.tex
Many applications of machine learning methods involve an iterative protocol in which data are collected, a model is trained, and then outputs of that model are used to choose what data to consider next.
For example, a data-driven approach for designing proteins is to train a regression model to predict the fitness of protein sequences, then use it to propose new sequences believed to exhibit greater fitness than observed in the training data.
Since validating designed sequences in the wet lab is typically costly, it is important to quantify the uncertainty in the model's predictions.
This is challenging because of a characteristic type of distribution shift between the training and test data that arises in the design setting---one in which the training and test data are statistically dependent, as the latter is chosen based on the former.
Consequently, the model's error on the test data---that is, the designed sequences---has an unknown and possibly complex relationship with its error on the training data.
We introduce a method to construct confidence sets for predictions in such settings, which account for the dependence between the training and test data.
The confidence sets we construct have finite-sample guarantees that hold for any prediction algorithm, even when a trained model chooses the test-time input distribution.
As a motivating use case, we use real data sets to demonstrate how our method quantifies uncertainty for the predicted fitness of designed proteins, and can therefore be used to select design algorithms that achieve acceptable trade-offs between high predicted fitness and low predictive uncertainty.

%% file: main.tex
\section{Uncertainty quantification under feedback loops}
\ifpnas
\dropcap{C}onsider a protein engineer who is interested in designing a protein with high \textit{fitness}---some real-valued measure of its desirability, such as fluorescence or therapeutic efficacy. The engineer has a data set of various protein sequences, denoted $X_i$, labeled with experimental measurements of their fitnesses, denoted $Y_i$, for $i = 1, \ldots, n$. The \emph{design problem} is to propose a novel sequence, $X_\text{test}$, that has higher fitness, $Y_\text{test}$, than any of these. To this end, the engineer trains a regression model on the data set, then identifies a novel sequence that the model predicts to be more fit than the training sequences. Can she trust the model's prediction for the designed sequence?
\else
Consider a protein engineer who is interested in designing a protein with high \textit{fitness}---some real-valued measure of its desirability, such as fluorescence or therapeutic efficacy. The engineer has a data set of various protein sequences, denoted $X_i$, labeled with experimental measurements of their fitnesses, denoted $Y_i$, for $i = 1, \ldots, n$. The \emph{design problem} is to propose a novel sequence, $X_\text{test}$, that has higher fitness, $Y_\text{test}$, than any of these. To this end, the engineer trains a regression model on the data set, then identifies a novel sequence that the model predicts to be more fit than the training sequences. Can she trust the model's prediction for the designed sequence? 
\fi

This is an important question to answer, not just for the protein design problem just described, but for any deployment of machine learning where the test data depends on the training data.
More broadly, settings ranging from Bayesian optimization to active learning to strategic classification involve \emph{feedback loops} in which the learned model and data influence each other in turn.
As feedback loops violate the standard assumptions of machine learning algorithms, we must be able to diagnose when a model's predictions can and cannot be trusted in their presence.

\begin{figure}
  \centering
  \includegraphics[width=8.7cm]{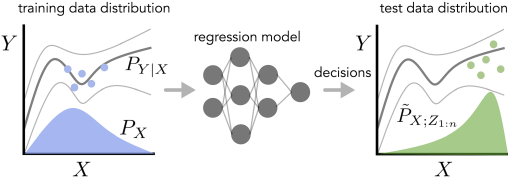}
  \caption{Illustration of feedback covariate shift. In the left graph, the blue distribution represents the training input distribution, $P_X$. The dark gray line sandwiched by lighter gray lines represents the mean $\pm$ the standard deviation of $P_{Y \mid X}$, the conditional distribution of the label given the input, which does not change between the training and test data distributions (left and right graphs, respectively). The blue dots represent training data, $Z_{1:n} = \{Z_1, \ldots, Z_n\}$, where $Z_i = (X_i, Y_i)$, which is used to fit a regression model (middle). Algorithms that use that trained model to make decisions, such as in design problems, active learning, and Bayesian optimization give rise to a new test-time input distribution, $P_{X; Z_{1:n}}$ (right graph, green distribution). The green dots represent test data.}
  \label{fig:acs}
\end{figure}

In this work, we address the problem of uncertainty quantification when the training and test data exhibit a type of dependence that we call \textit{feedback covariate shift} (FCS).
A joint distribution of training and test data falls under FCS if it satisfies two conditions.
First, the test input, $X_{\rm test}$, is selected based on independently and identically distributed (i.i.d.) training data,  $(X_1,Y_1), \ldots, (X_n,Y_n)$.
That is, the distribution of $X_{\rm test}$ is a function of the training data.
Second, $P_{Y \mid X}$, the ground-truth distribution of the label, $Y$, given any input, $X$, does not change between the training and test data distributions.
For example, returning to the example of protein design, the training data is used to select the designed protein, $X_{\rm test}$; the distribution of $X_{\rm test}$ is determined by some optimization algorithm that calls the regression model in order to design the protein.
However, since the fitness of any given sequence is some property dictated by nature, $P_{Y \mid X}$ stays fixed.
Representative examples of FCS include:
\begin{itemize}
    \item \textbf{Algorithms that use predictive models to explicitly choose the test distribution}, including the design of proteins, small molecules, and materials with favorable properties, active learning, adaptive experimental design, Bayesian optimization, and machine learning-guided scientific discovery. 
    
    \item \textbf{Algorithms that use predictive models to perform actions that change a system's state}, such as autonomous driving algorithms that use computer vision systems.
\end{itemize}
We anchor our discussion and experiments by focusing on protein design problems. However, the methods and insights developed herein are applicable to a variety of FCS problems.

\subsection{Quantifying uncertainty with valid confidence sets}

Given a regression model of interest, $\mu$, we quantify its uncertainty on an input with a \textit{confidence set}.
A confidence set is a function, $C: \mathcal{X} \rightarrow 2^\reals$, that maps a point from some input space, $\mathcal{X}$, to a set of real values that the model considers to be plausible labels.\footnote{We will use the term \textit{confidence set} to refer to both this function and the output of this function for a particular input; the distinction will be clear from the context.}
Informally, we will examine the model's error on the training data in order to quantify its uncertainty about the label, $Y_\text{test}$, of an input, $X_\text{test}$.
Formally, using the notation $Z_i = (X_i, Y_i), i = 1, \ldots, n$ and $Z_\text{test} = (X_\text{test}, Y_\text{test})$, our goal is to construct confidence sets that have the frequentist statistical property known as \textit{coverage}.
\begin{definition}
\label{def:valid}
Consider data points from some joint distribution, $(Z_1, \ldots, Z_n, Z_\textup{test}) \sim \mathcal{P}$. Given a \textup{miscoverage level}, $\alpha \in (0, 1)$, a confidence set, $C: \mathcal{X} \rightarrow 2^\reals$, which may depend on $Z_1, \ldots, Z_n$, provides \textup{coverage} under $\mathcal{P}$ if
\begin{align}
\label{eq:coverage}
    \Prob \left(Y_\textup{test} \in C(X_\textup{test}) \right) \geq 1 - \alpha,
\end{align}
where the probability is over all $n + 1$ data points, $(Z_1, \ldots, Z_n, Z_\textup{test}) \sim \mathcal{P}$.
\end{definition}

There are three important aspects of this definition. First, coverage is with respect to a particular joint distribution of the training and test data, $\mathcal{P}$, as the probability statement in \eqref{eq:coverage} is over random draws of all $n + 1$ data points.
That is, if one draws $(Z_1, \ldots, Z_n, Z_\text{test}) \sim \mathcal{P}$ and constructs the confidence set for $X_\textup{test}$ based on a regression model fit to $(Z_1, \ldots, Z_n)$, then the confidence set contains the true test label, $Y_\textup{test}$, a fraction of $1 - \alpha$ of the time. In this work, $\mathcal{P}$ can be any distribution captured by FCS, as we describe later in more detail.

Second, note that \eqref{eq:coverage} is a finite-sample statement: it holds for any number of training data points, $n$.
Finally, coverage is a marginal probability statement, which averages over all the randomness in the training and test data; it is not a statement about conditional probabilities, such as ${\Prob (Y_\textup{test} \in C(X_\textup{test}) \mid X_\textup{test})}$.
We will call a family of confidence sets, $C_\alpha$, indexed by the miscoverage level, $\alpha \in (0, 1)$, \emph{valid} if they provide coverage for all $\alpha \in (0, 1)$.

When the training and test data are exchangeable (e.g., independently and identically distributed), \textit{conformal prediction} is an approach for constructing valid confidence sets for any regression model \cite{vovk1999machine,vovk2005algorithmic,lei2018distribution}.
Though recent work has extended the methodology to certain forms of distribution shift \cite{tibshirani2019conformal, cauchois2020robust, gibbs2021adaptive, park2021pac, aleks2021distribution}, to our knowledge no existing approach can produce valid confidence sets when \textit{the test data depends on the training data}.
Here, we generalize conformal prediction to the FCS setting, enabling uncertainty quantification under this prevalent type of dependence between training and test data.

\subsection{Our contributions}

First, we formalize the concept of feedback covariate shift, which describes a type of distribution shift that emerges under feedback loops between learned models and the data they operate on.
Second, we introduce a generalization of conformal prediction that produces valid confidence sets under feedback covariate shift for any regression model.
We also introduce randomized versions of these confidence sets that achieve a stronger property called \emph{exact coverage}.
Finally, we demonstrate the use of our method to quantify uncertainty for the predicted fitness of designed proteins, using several real data sets.

We recommend using our method for design algorithm selection, as it enables practitioners to identify settings of algorithm hyperparameters that achieve acceptable trade-offs between high predictions and low predictive uncertainty.

\subsection{Prior work}

Our study investigates uncertainty quantification in a setting that brings together the well-studied concept of covariate shift~\cite{shimodaira2000, Sugiyama2005input, sugiyama2007covariate, quinonero2009dataset} with feedback between learned models and data distributions, a widespread phenomenon in real-world deployments of machine learning \cite{hardt2016strategic, perdomo2020performative}.
Indeed, beyond the design problem, feedback covariate shift is one way of describing and generalizing the dependence between data at successive iterations of active learning, adaptive experimental design, and Bayesian optimization.

Our work builds upon \textit{conformal prediction}, a framework for constructing confidence sets that satisfy the finite-sample coverage property in \eqref{eq:coverage} for arbitrary model classes \cite{gammerman1998learning, vovk2005algorithmic, angelopoulos2021gentle}.
Though originally based on the premise of exchangeable (e.g., independently and identically distributed) training and test data, the framework has since been generalized to handle various forms of distribution shift, including covariate shift \cite{tibshirani2019conformal, park2021pac}, label shift \cite{aleks2021distribution}, arbitrary distribution shifts in an online setting \cite{gibbs2021adaptive}, and test distributions that are nearby the training distribution \cite{cauchois2020robust}.
Conformal approaches have also been used to detect distribution shift \cite{vovk2020testing, hu2020distributionfree, luo2021sample, bates2021testing, angelopoulos2021learn, podkopaev2022tracking, kaur2022idecode}.

We call particular attention to the work of Tibshirani et al. \cite{tibshirani2019conformal} on conformal prediction in the context of covariate shift, whose technical machinery we adapt to generalize conformal prediction to feedback covariate shift.
In covariate shift, the training and test input distributions differ, but, critically, the training and test data are still independent; we henceforth refer to this setting as \textit{standard} covariate shift.
The chief innovation of our work is to formalize and address a ubiquitous type of dependence between training and test data that is absent from standard covariate shift, and, to the best of our knowledge, absent from any other form of distribution shift to which conformal approaches have been generalized.

For the design problem, in which a regression model is used to propose new inputs---such as a protein with desired properties---it is important to consider the predictive uncertainty of the designed inputs, so that we do not enter ``pathological'' regions of the input space where the model's predictions are desirable but untrustworthy~\cite{brookes2019cbas,fannjiang2020}.
Gaussian process regression (GPR) models are popular tools for addressing this issue, and algorithms that leverage their posterior predictive variance \cite{auer2002confidence, snoek2012bayesian} have been used to design enzymes with enhanced thermostability and catalytic activity \cite{romero2013gp, greenhalgh2021}.
Despite these successes, it is not clear how to obtain practically meaningful theoretical guarantees for the posterior predictive variance, and consequently to understand in what sense we can trust it. 
Similarly, ensembling strategies such as \cite{lakshminarayanan2017simple}, which are increasingly being used to quantify uncertainty for deep neural networks \cite{brookes2019cbas, zeng2019puffin, fannjiang2020, liu2020antibody}, as well as uncertainty estimates that are explicitly learned by deep models \cite{soleimany2021} do not come with formal guarantees.
A major advantage of conformal prediction is that it can be applied to any modelling strategy, and can be used to calibrate any existing uncertainty quantification approach, including those aforementioned.

\section{Conformal prediction under feedback covariate shift}

\subsection{Feedback covariate shift}
\label{sect:fcs}

We begin by formalizing feedback covariate shift (FCS), which describes a setting in which the test data depends on the training data, but the relationship between inputs and labels remains fixed.

We first set up our notation.
Recall that we let $Z_i = (X_i, Y_i)$, $i = 1, \ldots, n$, denote $n$ independently and identically distributed (i.i.d.) training data points comprising inputs, $X_i \in \mathcal{X}$, and labels, $Y_i \in \reals$.
Similarly, let $Z_\text{test} = (X_\text{test}, Y_\text{test})$ denote the test data point.
We use $Z_{1:n} = \{Z_1, \ldots, Z_n\}$ to denote the multiset of the training data, in which values are unordered but multiple instances of the same value appear according to their multiplicity.
We also use the shorthand $Z_{-i} = Z_{1:n} \setminus \{Z_i\}$, which is a multiset of $n - 1$ values that we refer to as the \textit{$i$-th leave-one-out training data set}. 

FCS describes a class of joint distributions over $(Z_1,\dots,Z_n,Z_\text{test})$ that have the dependency structure described informally in the Introduction.
Formally, we say that training and test data exhibit FCS when they can be generated according to the following three steps. 
\begin{enumerate}
    \item The training data, $(Z_1,\dots,Z_n)$, are drawn i.i.d.\ from some distribution:
\begin{align}
\begin{split}
X_i & \overset{\text{i.i.d}}\sim P_X, \\
Y_i & \sim P_{Y \mid X_i},\; i = 1, \ldots, n.
\end{split}
\end{align}

\item The realized training data induces a new input distribution over $\mathcal{X}$, denoted $\tilde{P}_{X; Z_{1:n}}$ to emphasize its dependence on the training data, $Z_{1:n}$.

\item The test input is drawn from this new input distribution, and its label is drawn from the unchanged conditional distribution:
\begin{align}
\begin{split}
    X_\text{test} & \sim \tilde{P}_{X; Z_{1:n}} \\
    Y_\text{test} & \sim P_{Y \mid X_\text{test}}.
\end{split}
\end{align}
\end{enumerate}

The key object in this formulation is the test input distribution, $\tilde{P}_{X; Z_{1:n}}$.
Prior to collecting the training data, $Z_{1:n}$, the specific test input distribution is not yet known.
The observed training data induces a distribution of test inputs, $\tilde{P}_{X; Z_{1:n}}$, that the model encounters at test time (for example, through any of the mechanisms summarized in the Introduction).

This is an expressive framework: the object $\tilde{P}_{X; Z_{1:n}}$ can be an arbitrarily complicated mapping from a data set of size $n$ to an input distribution, so long as it is invariant to the order of the data points.
There are no other constraints on this mapping; it need not exhibit any notion of smoothness, for example.
In particular, FCS encapsulates any design algorithm that makes use of a regression model fit to the training data, $Z_{1:n}$, in order to propose designed inputs.

\subsection{Conformal prediction for exchangeable data}

To explain how to construct valid confidence sets under FCS, we first walk through the intuition behind conformal prediction in the setting of exchangeable data, then present the adaptation to accommodate FCS.

\paragraph{Score function.}

First, we introduce the notion of a \textit{score function}, $S: (\mathcal{X} \times \reals) \times (\mathcal{X} \times \reals)^m \rightarrow \reals$, which is an engineering choice that quantifies how well a given data point ``conforms'' to a multiset of $m$ data points, in the sense of evaluating whether the data point comes from the same conditional distribution, $P_{Y \mid X}$, as the data points in the multiset.\footnote{Since the second argument is a multiset of data points, the score function must be invariant to the order of these data points. For example, when using the residual as the score, the regression model must be trained in a way that is agnostic to the order of the data points.}
A representative example is the residual score function, $S((X, Y), D) = |Y - \mu_D(X)|$, where $D$ is a multiset of $m$ data points and $\mu_D$ is a regression model trained on $D$.
A large residual signifies a data point that the model cannot easily predict, which suggests it does not obey the input-label relationship present in the training data.

More generally, we can choose the score to be any notion of uncertainty of a trained model on the point $(X, Y)$, heuristic or otherwise, such as the posterior predictive variance of a Gaussian process regression model \cite{romero2013gp, greenhalgh2021}, the variance of the predictions from an ensemble of neural networks \cite{lakshminarayanan2017simple, zeng2019puffin, liu2020antibody, angelopoulos2021gentle}, uncertainty estimates learned by deep models \cite{amini2020}, or even the outputs of other calibration procedures \cite{kuleshov2018accurate}.
Regardless of the choice of the score function, conformal prediction produces valid confidence sets; however, the particular choice of score function will determine the size, and therefore, informativeness, of the resulting sets. 
Roughly speaking, a score function that better reflects the likelihood of observing the given point, $(X, Y)$, under the true conditional distribution that governs $D$, $P_{Y \mid X}$, results in smaller valid confidence sets.

\paragraph{Imitating exchangeable scores.}
At a high level, conformal prediction is based on the observation that when the training and test data are exchangeable, their scores are also exchangeable.
More concretely, assume we use the residual score function, $S((X, Y), D) = |Y - \mu_D(X)|$, for some regression model class.
Now imagine that we know the label, $Y_\textup{test}$, for the test input, $X_\textup{test}$.
For each of the $n + 1$ training and test data points, $(Z_1, \ldots, Z_n, Z_\textup{test})$, we can compute the score using a regression model trained on the remaining $n$ data points; the resulting $n + 1$ scores are exchangeable.

In reality, of course, we do not know the true label of the test input.
However, this key property---that the scores of exchangeable data yield exchangeable scores---enables us to construct valid confidence sets by including all ``candidate'' values of the test label, $y \in \reals$, that yield scores for the $n + 1$ data points (the training data points along with the candidate test data point, $(X_\textup{test}, y)$) that appear to be exchangeable.
For a given candidate label, the conformal approach assesses whether or not this is true by comparing the score of the candidate test data point to an appropriately chosen quantile of the training data scores.

\subsection{Conformal prediction under FCS}

When the training and test data are under FCS, their scores are no longer exchangeable, since the training and test inputs are neither independent nor from the same distribution.
Our solution to this problem will be to weight each training and test data point to take into account these two factors.
Thereafter, we can proceed with the conformal approach of including all candidate labels such that the (weighted) candidate test data point is sufficiently similar to the (weighted) training data points.
Toward this end, we introduce two quantities: (1) a likelihood ratio function, which will be used to define the weights, and (2) the quantile of a distribution, which will be used to assess whether a candidate test data point conforms to the training data.

The likelihood ratio function for an input, $X$, which depends on a multiset of data points, $D$, is given by
\begin{align}
\label{eq:ratio}
v(X; D) &= \frac{\tilde{p}_{X; D}(X)}{p_X(X)},
\end{align}
where $\tilde{p}_{X; D}$ and $p_X$ denote the densities of the test and training input distributions, respectively, and the test input distribution is the particular one indexed by the data set, $D$.

This quantity is the ratio of the likelihoods under these two distributions, and as such, is reminiscent of weights used to modify various statistical procedures to accommodate standard covariate shift \cite{Sugiyama2005input, sugiyama2007covariate, tibshirani2019conformal}.
What distinguishes its use here is that our particular likelihood ratio is indexed by a multiset and depends on which data point is being evaluated as well as the candidate label, as will become clear shortly.

Consider a discrete distribution with probability masses $p_1, \ldots, p_n$ located at support points $s_1, \ldots, s_n$, respectively, where $s_i \in \reals$ and $p_i \geq 0, \sum_i p_i = 1$.
We define the \textit{$\beta$-quantile} of this distribution as
\begin{align}
\label{eq:quantile}
    \quantile{\beta} \left(\sum_{i = 1}^n p_i \, \delta_{s_i} \right) & = \inf \left\{s: \sum_{i: s_i \leq s} p_i \geq \beta \right\},
\end{align}
where $\delta_{s_i}$ is a unit point mass at $s_i$. 

We now define the confidence set.
For any score function, $S$, any miscoverage level, $\alpha \in (0, 1)$, and any test input, $X_\text{test} \in \mathcal{X}$, define the \emph{full conformal} confidence set as
\ifpnas
\begin{align}
\label{eq:confset}
    C_\alpha(X_\text{test}) = & \Bigg\{y \in \reals : S_{n + 1}(X_\text{test}, y) \leq \\
    & \quantile{1 - \alpha} \bigg( \sum_{i = 1}^{n + 1} w^y_i(X_\text{test}) \, \delta_{S_i(X_\text{test}, y)} \bigg) \Bigg\},
\end{align}
\else
\begin{align}
\label{eq:confset}
    C_\alpha(X_\text{test}) = & \Bigg\{y \in \reals : S_{n + 1}(X_\text{test}, y) \leq \quantile{1 - \alpha} \bigg( \sum_{i = 1}^{n + 1} w^y_i(X_\text{test}) \, \delta_{S_i(X_\text{test}, y)} \bigg) \Bigg\},
\end{align}
\fi
where 
\begin{align}
\begin{split}
    S_i(X_\text{test}, y) & = S(Z_i, Z_{-i} \cup \{(X_\text{test}, y)\}), \; i = 1, \ldots, n,\\
    S_{n + 1}(X_\textup{test}, y) & = S((X_\textup{test}, y), Z_{1:n}),
\end{split}
\end{align}
 which are the scores for each of the training and candidate test data points, when compared to the remaining $n$ data points, and the weights for these scores are given by
\begin{align}
\begin{split}
\label{eq:weights}
    w^y_i(X_\text{test}) & \propto  v(X_i; Z_{-i} \cup \{(X_\text{test}, y)\}), \; i = 1, \ldots, n, \\
    w^y_{n + 1}(X_\text{test}) & \propto  v(X_\text{test}; Z_{1:n}),
    %c & = \frac{1}{\sum_{j = 1}^{n} v(X_j; Z_{-j} \cup \{(X_\text{test}, y)\}) + v(X_\text{test}; Z_{1:n})}.
\end{split}
\end{align}
which are normalized such that $\sum_{i = 1}^{n + 1} w_i^y(X_\text{test}) = 1$.

In words, the confidence set in \eqref{eq:confset} includes all real values, $y \in \reals$, such that the ``candidate'' test data point, $(X_\text{test}, y)$, has a score that is sufficiently similar to the scores of the training data.
Specifically, the score of the candidate test data point needs to be smaller than the $(1-\alpha)$-quantile of the weighted scores of all $n + 1$ data points (the $n$ training data points as well as the candidate test data point), where the $i$-th data point is weighted by $w_i^y(X_\text{test})$.

Our main result is that this confidence set provides coverage under FCS (see \ifpnas SI Appendix S1.A \else Appendix \ref{app:proof} \fi for the proof).

\begin{theorem}
\label{thm}
Suppose data are generated under feedback covariate shift and assume $\tilde{P}_{X;D}$ is absolutely continuous with respect to $P_X$ for all possible values of $D$.
Then, for any miscoverage level, $\alpha \in (0, 1)$, the full conformal confidence set, $C_\alpha$, in \eqref{eq:confset} satisfies the coverage property in \eqref{eq:coverage}, namely, $\Prob(Y_\textup{test} \in C_\alpha(X_\textup{test}))\geq 1-\alpha$.
\end{theorem}

Since we can supply any domain-specific notion of uncertainty as the score function, this result implies we can interpret the condition in \eqref{eq:confset} as a calibration of the provided score function that guarantees coverage.
That is, our conformal approach can complement any existing uncertainty quantification method by endowing it with coverage under FCS.

We note that although Theorem \ref{thm} provides a lower bound on the probability $\Prob(Y_\textup{test} \in C_\alpha(X_\textup{test}))$, one cannot establish a corresponding upper bound without further assumptions on the training and test input distributions. 
However, by introducing randomization to the $\beta$-quantile, we can construct a randomized version of the confidence set, $C^\textup{rand}_\alpha(X_\textup{test})$, that is not conservative and satisfies ${\Prob(Y_\textup{test} \in C^\textup{rand}_\alpha(X_\textup{test})) = 1 - \alpha}$, a property called \emph{exact coverage}. See \ifpnas SI Appendix S1.B \else Theorem \ref{thm:random} \fi for details.

\paragraph{Estimating confidence sets in practice.}
In practice, one cannot check all possible candidate labels, $y \in \reals$, to construct a confidence set.
Instead, as done in previous work on conformal prediction, we estimate $C_\alpha(X_\textup{test})$ by defining a finite grid of candidate labels, $\mathcal{Y} \subset \reals$, and checking the condition in \eqref{eq:confset} for all $y \in \mathcal{Y}$.
Algorithm \ref{alg:generic} outlines a generic recipe for computing $C_\alpha(X_\text{test})$ for a given test input; see Section \ref{sec:ridge} for important special cases in which $C_\alpha(X_\text{test})$ can be computed more efficiently.

\begin{algorithm}[bth!]
\caption{Pseudocode for approximately computing $C_\alpha(X_\text{test})$}
\label{alg:generic}

\textbf{Input:} Training data, $(Z_1, \ldots, Z_n)$, where $Z_i = (X_i, Y_i)$;
test input, $X_\textup{test}$; finite grid of candidate labels, $\mathcal{Y} \subset \reals$; likelihood ratio function subroutine, $v(\cdot \, ; \cdot)$; and score function subroutine $S(\cdot, \cdot)$.

\textbf{Output:} Confidence set, $C_\alpha(X_\text{test}) \subset \mathcal{Y}$.

\begin{algorithmic}[1]
\State $C_\alpha(X_\text{test}) \gets \emptyset$

\State Compute $ v(X_\text{test}; Z_{1:n})$

\For{$y \in \mathcal{Y}$}

\For{$i = 1, \ldots, n$}
\State Compute $S_i(X_\text{test}, y)$ and ${v(X_i; Z_{-i} \cup \{(X_\text{test}, y)\})}$\;
\EndFor
\State Compute $S_{n + 1}(X_\text{test}, y)$\;

\For{$i = 1, \ldots, n + 1$}
\State Normalize $w_i^y(X_\text{test})$ according to \eqref{eq:weights}
\EndFor

\State $q_y \gets \quantile{1 - \alpha} \left( \sum_{i = 1}^{n + 1} w^y_i(X_\text{test}) \, \delta_{S_i(X_\text{test}, y)} \right)$

\If{$S_{n + 1}(X_\text{test}, y) \leq q_y$}
\State $C_\alpha(X_\text{test}) \gets C_\alpha(X_\text{test}) \cup \{y\}$
\EndIf
\EndFor
\end{algorithmic}

\end{algorithm}

\paragraph{Relationship with exchangeable and standard covariate shift settings.}

The weights assigned to each score, $w_i^y(X_\textup{test})$ in \eqref{eq:weights}, are the distinguishing factor between the confidence sets constructed by conformal approaches for the exchangeable, standard covariate shift, and FCS settings.
When the training and test data are exchangeable, these weights are simply $1/(n + 1)$.
To accommodate standard covariate shift, where the training and test data are independent, these weights are also normalized likelihood ratios---but, importantly, the test input distribution in the numerator is fixed, rather than data-dependent as in the FCS setting \cite{tibshirani2019conformal}.
That is, the weights are defined using one fixed likelihood ratio function, $v(\cdot) = \tilde{p}_X(\cdot) / p_X(\cdot)$, where $\tilde{p}_X$ is the density of the single test input distribution under consideration.

\sloppy
In contrast, under FCS, observe that the likelihood ratio that is evaluated in \eqref{eq:weights}, $v(\cdot; D)$ from \eqref{eq:ratio}, is different for each of the $n + 1$ training and candidate test data points and for each candidate label, $y \in \reals$.
To weight the $i$-th training score, we evaluate the likelihood ratio of $X_i$ where the test input distribution is the one induced by ${Z_{-i} \cup \{(X_\textup{test}, y)\}}$,
\begin{align}
    v(X_i; Z_{-i} \cup \{(X_\textup{test}, y)\}) = \frac{\tilde{p}_{X; Z_{-i} \cup \{(X_\textup{test}, y)\}}(X_i)}{p_X(X_i)}.
\end{align}
That is, the weights under FCS take into account not just a single test input distribution, but every test input distribution that can be induced when we treat a leave-one-out training data set combined with a candidate test data point, ${Z_{-i} \cup \{(X_\textup{test}, y)\}}$, as the training data.

To further appreciate the relationship between the standard and feedback covariate shift settings, consider the weights used in the standard covariate shift approach if we treat $P_{X; Z_{1:n}}$ as the test input distribution.
The extent to which $P_{X; Z_{1:n}}$ differs from $P_{X; Z_{-i} \cup \{(X_\textup{test}, y)\}}$, for any $i = 1, \ldots, n$ and $y \in \reals$, determines the extent to which the weights used under standard covariate shift deviate from those used under FCS.
In other words, since $Z_{1 : n}$ and $Z_{-i} \cup \{(X_\textup{test}, y\}$ differ in exactly one data point, the similarity between the standard covariate shift and FCS weights depends on the ``smoothness'' of the mapping from $D$ to $\tilde{P}_{X; D}$.
For example, the more algorithmically stable the learning algorithm through which $\tilde{P}_{X; D}$ depends on $D$ is, the more similar these weights will be.

\paragraph{Input distributions are known in the design problem.}

The design problem is a unique setting in which we have control over the data-dependent test input distribution, $P_{X; D}$, since we choose the procedure used to design an input.
In the simplest case, some design procedures sample from a distribution whose form is explicitly chosen, such as an energy-based model whose energy function is proportional to the predictions from a trained regression model \cite{biswas2020low}, or a model whose parameters are set by solving an optimization problem (e.g., the training of a generative model) \cite{popova2018, kang2019, brookes2019cbas, fannjiang2020, russ2020, Wu2020-lb, Hawkins-Hooker2021-nz, Shin2021-xg, zhu2021aav}.
In either setting, we know the exact form of the test input distribution, which also absolves the need for density estimation.

In other cases, the design procedure involves iteratively applying a gradient to, or otherwise locally modifying, an initial input in order to produce a designed input \cite{killoran2017, gomezbombarelli2018, linder2020, Sinai2020-nv, bashir2021, Bryant2021-yj}.
Due to randomness in either the initial input or the local modification rule, such procedures implicitly result in some distribution of test inputs.
Though we do not have access to its explicit form, knowledge of the design procedure can enable us to estimate it much more readily than in a naive density estimation setting.
For example, we can simulate the design procedure as many times as is needed to sufficiently estimate the resulting density, whereas in density estimation in general, we cannot control how many test inputs we can access.

The training input distribution, $P_X$, is also often known explicitly.
In protein design problems, for example, training sequences are often generated by introducing random substitutions to a single wild type sequence \cite{brookes2019cbas,biswas2020low, Bryant2021-yj}, by recombining segments of several ``parent'' sequences \cite{li2007cytochrome, romero2013gp, bedbrook2019, greenhalgh2021}, or by independently sampling the amino acid at each position from a known distribution \cite{zhu2021aav, Weinstein2021-dc}.
Conveniently, we can then compute the weights in \eqref{eq:weights} exactly without introducing approximation error due to density ratio estimation.

Finally, we note that, by construction, the design problem tends to result in test input distributions that place considerable probability mass on regions where the training input distribution does not.
The further the test distribution is from the training distribution in this regard, the larger the resulting weights on candidate test points, and the larger the confidence set in \eqref{eq:confset} will tend to be.
This phenomenon agrees with our intuition about epistemic uncertainty: we should have more uncertainty---that is, larger confidence sets---in regions of input space where there is less training data.

\subsection{Efficient computation of confidence sets under feedback covariate shift}
\label{sec:ridge}

Using  Algorithm \ref{alg:generic} to construct the full conformal confidence set, $C_\alpha(X_\text{test})$, requires computing the scores and weights, $S_i(X_\text{test}, y)$ and $w_i^y(X_\text{test})$, for all $i = 1, \ldots, n + 1$ and all candidate labels, $y \in \mathcal{Y}$.
When the dependence of $\tilde{P}_{X; D}$ on $D$ arises from a model trained on $D$, then naively, we must train $(n + 1) \times |\mathcal{Y}|$ models in order to compute these quantities.
We now describe two important, practical cases in which this computational burden can be reduced to fitting $n + 1$ models, removing the dependence on the number of candidate labels.
In such cases, we can post-process the outputs of these $n + 1$ models to calculate all $(n + 1) \times |\mathcal{Y}|$ required scores and weights (see \ifpnas Algorithm S2 in the SI Appendix \else Alg. \ref{alg:ridge} in the Appendix \fi for pseudocode); we refer to this as computing the confidence set efficiently.

In the following two examples and in our experiments, we use the residual score function, $S((X, Y), D) = |Y - \mu_D(X)|$, where $\mu_D$ is a regression model trained on the multiset $D$.
To understand at a high level when efficient computation is possible, first let $\augloomu{i}$ denote the regression model trained on $\augloodata{i} = Z_{-i} \cup \{(X_\text{test}, y)\}$, the $i$-th leave-one-out training data set combined with a candidate test data point.
The scores and weights can be computed efficiently when $\augloomu{i}(X_i)$ is a computationally simple function of the candidate label, $y$, for all $i$---for example, a linear function of $y$.
We discuss two such cases in detail.

\paragraph{Ridge regression.}

Suppose we fit a ridge regression model, with ridge regularization hyperparameter $\gamma$, to the training data.
Then, we draw the test input vector from a distribution which places more mass on regions of $\mathcal{X}$ where the model predicts more desirable values, such as high fitness in protein design problems.
Recent studies have employed this relatively simple approach to successfully design novel enzymes with enhanced catalytic efficiencies and thermostabilities \cite{biswas2020low, li2007cytochrome, fox2007prosar}.

In the ridge regression setting, the quantity $\augloomu{i}(X_i)$ can be written in closed form as
\begin{align}
\label{eq:ridgepred}
    \augloomu{i}(X_i) & = \left[\left(\mathbf{X}_{-i}^T\mathbf{X}_{-i} + \gamma I \right)^{-1} \mathbf{X}_{-i}^T Y^y_{-i} \right]^T X_i \\
    & = \left(\sum_{j = 1}^{n - 1} Y_{-i;j} \mathbf{A}_{-i; j}\right)^T X_i + (\mathbf{A}_{-i; n}^T X_i) y,
\end{align}
where the rows of the matrix $\mathbf{X}_{-i} \in \reals^{n \times p}$ are the input vectors in $\augloodata{i}$, $Y^y_{-i} = (Y_{-i}, y)\in \reals^n$ contains the labels in $\augloodata{i}$, the matrix $\mathbf{A}_{-i} \in \reals^{n \times p}$ is defined as ${\mathbf{A}_{-i} = \left(\mathbf{X}_{-i}^T\mathbf{X}_{-i} + \gamma I \right)^{-1} \mathbf{X}_{-i}^T}$, $\mathbf{A}_{-i; j}$ denotes the $j$-th column of $A_{-i}$, and $Y_{-i;j}$ denotes the $j$-th element of $Y_{-i}$.

Note that the expression in \eqref{eq:ridgepred} is a linear function of the candidate label, $y$.
Consequently, as instantiated in \ifpnas Alg. S2 in the SI Appendix, \else Alg. \ref{alg:ridge} in the Appendix\fi, we first compute and store the slopes and intercepts of these linear functions for all $i$, which can be calculated as byproducts of fitting $n + 1$ ridge regression models.
Using these parameters, we can then compute $\augloomu{i}(X_i)$ for all candidate labels, $y \in \mathcal{Y}$, by simply evaluating a linear function of $y$ instead of retraining a regression model on $\augloodata{i}$.
Altogether, beyond fitting $n + 1$ ridge regression models, \ifpnas Alg. S2 \else Alg. \ref{alg:ridge} in the Appendix \fi requires $O(n \cdot p \cdot |\mathcal{Y}|)$ additional floating point operations to compute the scores and weights for all the candidate labels, the bulk of which can be implemented as one outer product between an $n$-vector and a $|\mathcal{Y}|$-vector, and one Kronecker product between an $(n \times p)$-matrix and a $|\mathcal{Y}|$-vector.

\paragraph{Gaussian process regression.}

Similarly, suppose we fit a Gaussian process regression model to the training data.
We then select a test input vector according to a likelihood that is a function of the mean and variance of the model's prediction; such functions are referred to as \emph{acquisition functions} in the Bayesian optimization literature.

For a linear kernel, the expression for the mean prediction, $\augloomu{i}(X_i)$, is the same as for ridge regression (\eqref{eq:ridgepred}). For arbitrary kernels, the expression can be generalized and remains a linear function of $y$ (see \ifpnas SI Appendix S2.B \else Appendix \ref{app:gpr} \fi for details).
We can therefore mimic the computations described for the ridge regression case to compute the scores and weights efficiently.

\subsection{Data splitting}
\label{sect:split}

For settings with abundant training data, or model classes that do not afford efficient computations of the scores and weights, one can turn to \emph{data splitting} to construct valid confidence sets.
To do so, we first randomly partition the labeled data into disjoint training and calibration sets.
Next, we use the training data to fit a regression model, which induces a test input distribution.
If we condition on the training data, thereby treating the regression model as fixed, we have a setting in which (1) the calibration and test data are drawn from different input distributions, but (2) are independent (even though the test and training data are not).
Thus, data splitting returns us to the setting of standard covariate shift, under which we can use the data splitting approach in \cite{tibshirani2019conformal} to construct valid \emph{split conformal} confidence intervals (see \ifpnas SI Appendix S1.C). \else Section \ref{app:split}). \fi

We also introduce randomized data splitting approaches that give exact coverage; see \ifpnas SI Appendix S1.D \else Appendix \ref{app:rand-split} \fi for details.

\section{Experiments with protein design}

To demonstrate practical applications of our work, we turn to examples of uncertainty quantification for designed proteins.
Given a fitness function\footnote{We use the term \emph{fitness function} to refer to a particular property that can be exhibited by proteins, while the \emph{fitness} of a protein refers to the extent to which it exhibits that property.} of interest, such as fluorescence, a typical goal of protein design is to seek a protein with high fitness---in particular, higher than we have observed in known proteins.
Historically, this has been accomplished in the wet lab through several iterations of expensive, time-consuming experiments.
Recently, efforts have been made to augment such approaches with machine learning-based strategies; see reviews by Yang et al. \cite{Yang2019-ac}, Sinai \& Kelsic \cite{Sinai2020-qy}, and Wu et al. \cite{Wu2021-si} and references therein.
For example, one might train a regression model on protein sequences with experimentally measured fitnesses, then use an optimization algorithm or fit a generative model that leverages that regression model to propose promising new proteins \cite{fox2007prosar,brookes2019cbas, romero2013gp,bedbrook2019,wu2019mlde,Angermueller2019-iclr,biswas2020low,linder2020,greenhalgh2021,wittmann2021mlde, zhu2021aav}.
Special attention has been given to the \textit{single-shot} case in which we are given just a single batch of training data, due to its obvious practical convenience.

The use of regression models for design involves balancing (1) the desire to explore regions of input space far from the training inputs, in order to find new desirable inputs, with (2) the need to stay close enough to the training inputs that we can trust the regression model.
As such, estimating predictive uncertainty in this setting is important.
Furthermore, the training and designed data are described by feedback covariate shift: since the fitness is some quantity dictated by nature, the conditional distribution of fitness given any sequence stays fixed, but the distribution of designed sequences is chosen based on a trained regression model.\footnote{In this section, we will use ``test'' and ``designed'' interchangeably when describing data. We will also sometimes say ``sequence'' instead of ``input,'' but this does not imply any constraints on how the protein is represented or featurized.}

\begin{figure}
  \centering
  \includegraphics[width=8cm]{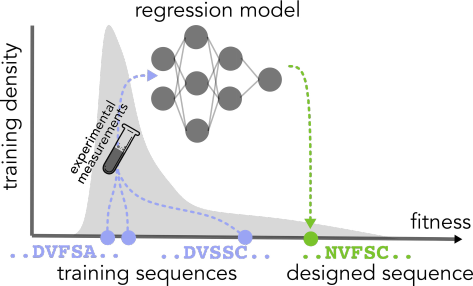}  % 8.7cm
  \caption{Illustration of single-shot protein design. The gray distribution represents the distribution of fitnesses under the training sequence distribution. The blue circles represent the fitnesses of three training sequences, and the goal is to propose a sequence with even higher fitness. To that end, we fit a regression model to the training sequences labeled with experimental measurements of their fitnesses, then deploy some design procedure that uses that trained model to propose a new sequence believed to have a higher fitness (green circle).}
  \label{fig:protein}
\end{figure}

Our experimental protocol is as follows.
Given training data consisting of protein sequences labeled with experimental measurements of their fitnesses, we fit a regression model, then sample test sequences (representing designed proteins) according to design algorithms used in recent work \cite{biswas2020low, zhu2021aav} (Fig.~\ref{fig:protein}).
We then construct confidence sets with guaranteed coverage for the designed proteins, and examine various characteristics of those sets to evaluate the utility of our approach.
In particular, we show how our method can be used to select design algorithm hyperparameters that achieve acceptable trade-offs between high predicted fitness and low predictive uncertainty for the designed proteins.
Code reproducing these experiments is available at \texttt{\href{https://github.com/clarafy/conformal-for-design}{https://github.com/clarafy/conformal-for-design}}.

\subsection{Design experiments using combinatorially complete fluorescence data sets}
\label{sect:protein-design}

The challenge when evaluating \textit{in silico} design methods is that in general, we do not have labels for the designed sequences.
One workaround, which we take here, is to make use of \emph{combinatorially complete} protein data sets~\cite{poelwijk2019, wu2019mlde,wittmann2021mlde,brookes2022sparsity}, in which a small number of fixed positions are selected from some wild type sequence, and all possible variants of the wild type that vary in those selected positions are measured experimentally.
Such data sets enable us to simulate protein design problems where we always have labels for the designed sequences.
In particular, we can use a small subset of the data for training, then deploy a design procedure that proposes novel proteins (restricted to being variants of the wild type at the selected positions), for which we have labels.

We used data of this kind from Poelwijk et al.\ \cite{poelwijk2019}, which focused on two ``parent'' fluorescent proteins that differ at exactly thirteen positions in their sequences, and are identical at every other position.
All $2^{13} = 8,192$ sequences that have the amino acid of either parent at those thirteen sites (and whose remaining positions are identical to the parents) were experimentally labeled with a measurement of brightness at both a ``red'' wavelength and a ``blue'' wavelength, resulting in combinatorially complete data sets for two different fitness functions.
In particular, for both wavelengths, the label for each sequence was an enrichment score based on the ratio of its counts before and after brightness-based selection through fluorescence-activated cell sorting.
The enrichment scores were then normalized so that the same score reflects comparable brightness for both wavelengths.

Finally, each time we sampled from this data set to acquire training or designed data, as described below, we added simulated measurement noise to each label by sampling from a noise distribution estimated from the combinatorially complete data set (see \ifpnas S3 in the SI Appendix for details).\else \ref{app:protein-results} for details).\fi
This step simulates the fact that sampling and measuring the same sequence multiple times results in different measurements.

\subsubsection{Protocol for design experiments}

Our training data sets consisted of $n$ data points, $Z_{1:n}$, sampled uniformly at random from the combinatorially complete data set.
We used $n \in \{96, 192, 384 \}$ as is typical of realistic scenarios~\cite{romero2013gp,bedbrook2019,wu2019mlde,biswas2020low,wittmann2021mlde}.
We represented each sequence as a feature vector containing all first- and second-order interaction terms between the thirteen variable sites, and fit a ridge regression model, $\mu_{Z_{1:n}}(x)$, to the training data, where the regularization strength was set to $10$ for $n = 96$ and $1$ otherwise.
Linear models of interaction terms between sequence positions have been observed to be both theoretically justified and empirically useful as models of protein fitness functions \cite{poelwijk2019, hsu2022, brookes2022sparsity} and thus may be particularly useful for protein design, particularly with small amounts of training data.

\paragraph{Sampling designed sequences.}
Following ideas in \cite{biswas2020low, zhu2021aav}, we designed a protein by sampling from a sequence distribution whose log-likelihood is proportional to the prediction of the regression model:
\begin{align}
\label{eq:protein-test-dist}
    \tilde{p}_{X; Z_{1:n}}(X_\text{test}) \propto \exp(\lambda \cdot \mu_{Z_{1:n}}(X_\text{test})),
\end{align}
where $\lambda > 0$, the \textit{inverse temperature}, is a hyperparameter.
Larger values of $\lambda$ result in distributions of designed sequences that are more likely to have high predicted fitnesses according to the model, but are also, for this same reason, more likely to be in regions of sequence space that are further from the training data and over which the model is more uncertain.
Analogous hyperparameters have been used in recent protein design work to control this trade-off between exploration and exploitation \cite{biswas2020low, russ2020, Madani2021-le, zhu2021aav}.
We took $\lambda \in \{0, 2, 4, 6\}$ to investigate how the behavior of our confidence sets varies along this trade-off. 

\paragraph{Constructing confidence sets for designed sequences.}
For each setting of $n$ and $\lambda$, we generated $n$ training data points and one designed data point as just described $T = 2000$ times.
For each of these $T$ trials, we used \ifpnas Alg.\ S2 in the SI Appendix \else Alg. \ref{alg:ridge} in the Appendix \fi to construct the full conformal confidence set, $C_\alpha(X_\text{test})$, using a grid of real values between $0$ and $2.2$ spaced $\Delta = 0.02$ apart as the set of candidate labels, $\mathcal{Y}$.
This range contained the ranges of fitnesses in both the blue and red combinatorially complete data sets, $[0.091, 1.608]$ and $[0.025, 1.692]$, respectively.\footnote{In general, a reasonable approach for constructing a finite grid of candidate labels, $\mathcal{Y}$, is to span an interval beyond which one knows label values are impossible in practice, based on prior knowledge about the measurement technology.
The presence or absence of any such value in a confidence set would not be informative to a practitioner.
The size of the grid spacing, $\Delta$, determines the resolution at which we evaluate coverage; that is, in terms of coverage, including a candidate label is equivalent to including the $\Delta$-width interval centered at that label value.
Generally, one should therefore set $\Delta$ as small as possible, subject to one's computational budget.}

We used $\alpha = 0.1$ as a representative miscoverage value, corresponding to coverage of $1 - \alpha = 0.9$.
We then computed the \textit{empirical coverage} achieved by the confidence sets, defined as the fraction of the $T$ trials where the true fitness of the designed protein was within half a grid spacing from some value in the confidence set, namely, $\min\{|Y_\text{test} - y| : y \in C_\alpha(X_\text{test})\} \leq \Delta / 2$.
Based on Theorem \ref{thm}, assuming $\mathcal{Y}$ is both a large and fine enough grid to encompass all possible fitness values, the expected empirical coverage is lower bounded by $1 - \alpha = 0.9$.
However, there is no corresponding upper bound, so it will be of interest to examine any excess in the empirical coverage, which corresponds to the confidence sets being conservative (larger than necessary).
Ideally, the empirical coverage is exactly $0.9$, in which case the sizes of the confidence sets reflect the minimal predictive uncertainty we can have about the designed proteins while achieving coverage.

In our experiments, the computed confidence sets tended to comprise grid-adjacent candidate labels, suggestive of confidence intervals.
As such, we hereafter refer to the \textit{width} of \textit{confidence intervals}, defined as the grid spacing size times the number of values in the confidence set, $\Delta \cdot  |C_\alpha(X_\text{test})|$.

\subsubsection{Results}
\label{sect:fluorescence-results}

Here we discuss results for the blue fluorescence data set. Analogous results for the red fluorescence data set are presented in \ifpnas SI Appendix S3. \else Appendix \ref{app:protein-results}. \fi

\begin{figure*}
  \centering
  \includegraphics[width=17.8cm]{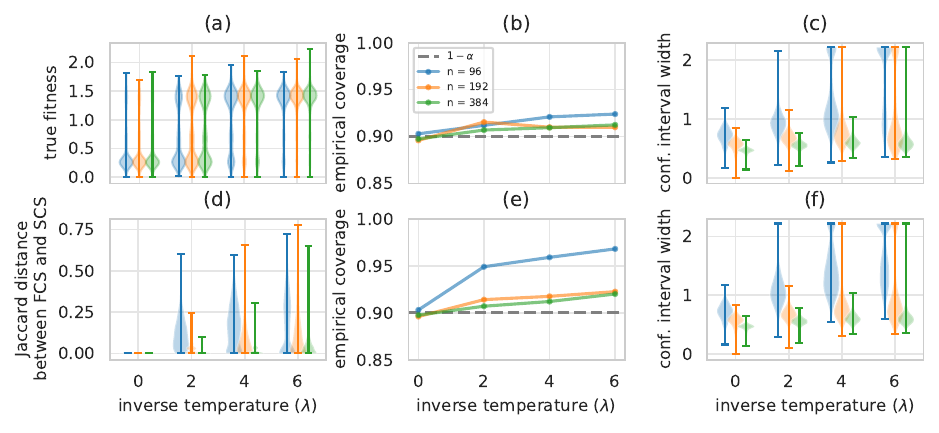}  % 17.8cm
  \caption{Quantifying predictive uncertainty for designed proteins, using the blue fluorescence data set. (a) Distributions of labels of designed proteins, for different values of the inverse temperature, $\lambda$, and different amounts of training data, $n$. Labels surpass the fitness range observed in the combinatorially complete data set, $[0.091, 1.608]$, due to additional simulated measurement noise. (b) Empirical coverage, compared to the theoretical lower bound of $1 - \alpha = 0.9$ (dashed gray line), and (c) distributions of confidence interval widths achieved by full conformal prediction for feedback covariate shift (our method) over $T = 2000$ trials. In (a), and (c), the whiskers signify the minimum and maximum observed values. (d) Distributions of Jaccard distances between the confidence intervals produced by full conformal prediction for feedback covariate shift and standard covariate shift \cite{tibshirani2019conformal}. (e, f) Same as (b, c) but using full conformal prediction for standard covariate shift.}
  \label{fig:blue-marginal}
\end{figure*}

\paragraph{Effect of inverse temperature.}
First we examined the effect of the inverse temperature, $\lambda$, on the fitnesses of designed proteins (Fig.~\ref{fig:blue-marginal}a).
Note that $\lambda = 0$ corresponds to a uniform distribution over all sequences in the combinatorially complete data set (i.e., the training distribution), which mostly yields label values less than $0.5$.
For $\lambda \geq 4$, we observe a considerable mass of designed proteins attaining fitnesses around $1.5$, so these values of $\lambda$ represent settings where the designed proteins are more likely to be fitter than the training proteins.
This observation is consistent with the use of this and other analogous hyperparameters to tune the outcomes of design algorithms~\cite{russ2020,biswas2020low, Madani2021-le,zhu2021aav}, and is meant to provide an intuitive interpretation of the hyperparameter to readers unfamiliar with its use in design problems.

\paragraph{Empirical coverage and confidence interval widths.}

Despite the lack of a theoretical upper bound, the empirical coverage does not tend to exceed the theoretical lower bound of $1 - \alpha = 0.9$ by much (Fig.~\ref{fig:blue-marginal}b), reaching at most $0.924$ for $n = 96, \lambda = 6$.
Loosely speaking, this observation suggests that the confidence intervals are nearly as small, and therefore as informative, as they can be while achieving coverage.

As for the widths of the confidence intervals, we observe that for any value of $\lambda$, the intervals tend to be smaller for larger amounts of training data (Fig.~\ref{fig:blue-marginal}c).
Also, for any value of $n$, the intervals tend to get larger as $\lambda$ increases. 
The first phenomenon agrees with the intuition that training a model on more data should generally reduce predictive uncertainty.
The second phenomenon arises because greater values of $\lambda$ lead to designed sequences with higher predicted fitnesses, which the model is more uncertain about.
Indeed, for $\lambda = 4, n = 96$ and $\lambda = 6, n \in \{96, 192\}$, many confidence intervals contain the entire range of fitnesses in the combinatorially complete data set.
In these regimes, the regression model cannot glean enough information from the training data to have much certainty about the designed protein.

\paragraph{Comparison to standard covariate shift}
 
Deploying full conformal prediction as prescribed for standard covariate shift (SCS) \cite{tibshirani2019conformal}, a heuristic with no formal guarantees in this setting, often results in more conservative confidence sets than those produced by our method (Fig.~\ref{fig:blue-marginal}).
To understand when the outputs of these two methods will differ more or less, we can compare the forms of the weights that both methods introduce on the training and candidate test data points when considering a candidate label.

First, recall that for both feedback covariate shift (FCS) and SCS, the weight assigned to the $i$-th training score is a normalized ratio of the likelihood of $X_i$ under a test input distribution and the training input distribution, $p_X$, namely:
\begin{align}
\begin{split}
\label{eq:scs-vs-fcs}
    v(X_i; Z_{-i} \cup \{(X_\textup{test}, y)\}) & = \tilde{p}_{X; Z_{-i} \cup \{(X_\textup{test}, y)\}} (X_i) / p_X(X_i), \\
    v(X_i) & = \tilde{p}_{X; Z_{1:n}}(X_i) / p_X(X_i),
\end{split}
\end{align}
for FCS and SCS, respectively.
For FCS, the test input distribution, $\tilde{p}_{X; Z_{-i} \cup \{(X_\textup{test}, y)\}}$, is induced by a regression model trained on $Z_{-i} \cup \{(X_\textup{test}, y)\}$, and therefore depends on the candidate label, $y$, and also differs for each of the $n$ training inputs.
Consequently, for FCS the weight on the $i$-th training score depends on the candidate label under consideration, $y$.
In contrast, for SCS the test input distribution, $p_{X; Z_{1:n}}$, is simply the one induced by the training data, $Z_{1:n}$, and is therefore fixed for all training scores and all candidate labels.

Note, however, that the SCS and FCS weights depend on data sets, $Z_{1:n}$ and $Z_{-i} \cup \{(X_\textup{test}, y)\}$, respectively, that differ only in a single data point: the former contains $Z_i$, while the latter contains $(X_\textup{test}, y)$.
Therefore, the difference between the weights---and the resulting confidence sets---is a direct consequence of how sensitive the mapping from data set to test input distribution, $D \rightarrow \tilde{p}_{X; D}$ (given by \eqref{eq:protein-test-dist} in this setting), is to changes of a single data point in $D$.
Roughly speaking, the less sensitive this mapping, the more similar the FCS and SCS confidence sets will be.
For example, using more training data (e.g., $n = 384$ compared to $n = 96$ for a fixed $\lambda$) or a lower inverse temperature (e.g., $\lambda = 2$ compared to $\lambda = 6$ for a fixed $n$) results in more similar SCS and FCS confidence sets (Figs.~\ref{fig:blue-marginal}d, \ifpnas S2d, S5\else \ref{fig:red-marginal},~\ref{fig:weights-comparison-lambda}\fi).
Similarly, using regression models with fewer features or stronger regularization also results in more similar confidence sets (\ifpnas Figs. S3, S4, S6\else Figs.~\ref{fig:higher-orders},~\ref{fig:regs},~\ref{fig:weights-comparison-reg}\fi).

One can therefore think of and use SCS confidence sets as a computationally cheaper approximation to FCS confidence sets, where the approximation is better for mappings ${D \rightarrow \tilde{p}_{X; D}}$ that are less sensitive to changes in $D$.
Conversely, the extent to which SCS confidence sets are similar to FCS confidence sets will generally reflect this sensitivity.
In our protein design experiments, SCS confidence sets tend to be more conservative than their FCS counterparts, where the extent of overcoverage generally increases with less training data, higher inverse temperature (Figs. \ref{fig:blue-marginal}d, \ifpnas S2d\else \ref{fig:red-marginal}\fi), more complex features, and weaker regularization (\ifpnas Figs. S3, S4\else Figs.~\ref{fig:higher-orders},~\ref{fig:regs}\fi). 

\paragraph{Using uncertainty quantification to set design procedure hyperparameters.}

\begin{figure*}
  \centering
  \includegraphics[width=17.8cm]{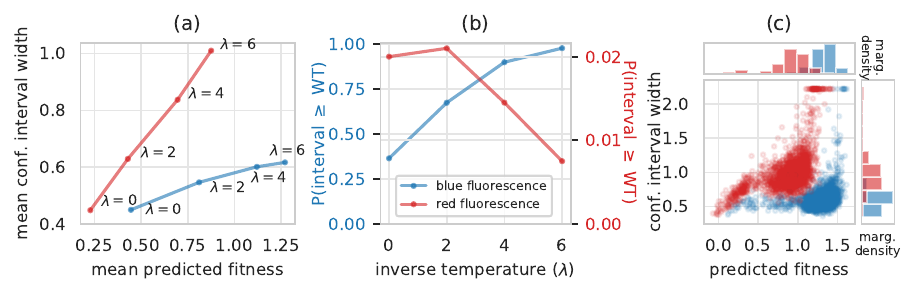}
  \caption{Comparison of trade-off between predicted fitness and predictive certainty on the red and blue fluorescence data sets. (a) Trade-off between mean confidence interval width and mean predicted fitness for different values of the inverse temperature, $\lambda$, and $n = 384$ training data points. (b) Empirical probability that the smallest fitness value in the confidence intervals of designed proteins exceeds the true fitness of one of the wild-type parent sequences, mKate2. (c) For $n = 384$ and $\lambda = 6$, the distributions of both confidence interval width and predicted fitnesses of designed proteins.}
  \label{fig:redvblue}
\end{figure*}

As the inverse temperature, $\lambda$, in \eqref{eq:protein-test-dist} varies, there is a trade-off between the mean predicted fitness and predictive certainty for designed proteins: both mean predicted fitness and mean confidence interval width grow as $\lambda$ increases (Fig.~\ref{fig:redvblue}a).
To demonstrate how our method might be used to inform the design procedure itself, one can visualize this trade-off (Fig.~\ref{fig:redvblue}) and use it to decide on a setting of $\lambda$ that achieves both a mean predicted fitness and degree of certainty that one finds acceptable, given, for example, some resource budget for evaluating designed proteins in the wet lab.
For data sets of different fitness functions, which may be better or worse approximated by our chosen regression model class and may have different amounts of measurement noise, this trade-off---and therefore, the appropriate setting of $\lambda$---will be different (Fig.~\ref{fig:redvblue}).

For example, protein design experiments on the red fluorescence data set result in a less favorable trade-off between mean predicted fitness and predictive certainty than the blue fluorescence data set: the same amount of increase in mean predicted fitness corresponds to a greater increase in mean interval width for red compared to blue fluorescence (Fig.~\ref{fig:redvblue}a).
We might therefore choose a smaller value of $\lambda$ when designing proteins for the former compared to the latter.
Indeed, predictive uncertainty grows so quickly for red fluorescence that, for $\lambda > 2$, the empirical probability that the smallest value in the confidence interval is greater than the true fitness of a wild type sequence decreases rather than increases (Fig.~\ref{fig:redvblue}b) which suggests we may not want to set $\lambda > 2$.
In contrast, if we had looked at the mean predicted fitness alone without assessing the uncertainty of those predictions, it grows monotonically with $\lambda$ (Fig.~\ref{fig:redvblue}a), which would not suggest any harm from setting $\lambda$ to a higher value.

In contrast, for blue fluorescence, although the mean interval width also grows with $\lambda$, it does so at a much slower rate than for red fluorescence (Fig.~\ref{fig:redvblue}a); correspondingly, the empirical frequency at which the confidence interval surpasses the fitness of the wild type also grows monotonically (Fig.~\ref{fig:redvblue}b).

We can observe these differences in the trade-off between blue and red fluorescence even for a fixed value of $\lambda$.
For example, for $n = 384, \lambda = 6$ (Fig.~\ref{fig:redvblue}c), observe that proteins designed for blue fluorescence (blue circles) mostly lie in a flat horizontal band.
That is, those with higher predicted fitnesses do not have much wider intervals than those with lower predicted fitnesses, except for a few proteins with the highest predicted fitnesses.
In contrast, for red fluorescence, designed proteins with higher predicted fitnesses also tend to have wider confidence intervals.

\subsection{Design experiments using adeno-associated virus (AAV) capsid packaging data}

In contrast with Section \ref{sect:protein-design}, which represented a protein design problem with limited amounts of labeled data (at most a few hundred sequences), here we focus on a setting in which there is abundant labeled data.
We can therefore employ data splitting as described in Section \ref{sect:split} to construct confidence sets, as an alternative to computing full conformal confidence sets (\eqref{eq:confset}) as done in Section \ref{sect:protein-design}.
Specifically, we construct a randomized version of the split conformal confidence set (\ifpnas S1.D in the SI Appendix\else Section \ref{app:rand-split}\fi), which achieves exact coverage.

This subsection, together with the previous subsection, demonstrate that in both regimes---limited and abundant labeled data---our proposed methods provide confidence sets that give coverage, are not overly conservative, and can be used to visualize the trade-off between predicted fitness and predictive uncertainty inherent to a design algorithm.

\subsubsection{Protein design problem: AAV capsid proteins with improved packaging ability}

Adeno-associated viruses (AAVs) are a class of viruses whose capsid, the protein shell that encapsulates the viral genome, are a promising delivery vehicle for gene therapy.
As such, the proteins that constitute the capsid have been modified to enhance various fitness functions, such as the ability to enter specific cell types and evade the immune system \cite{Maheshri2006-bg,Dalkara2013-wy, Tse2017-ga}.
Such efforts usually start by sampling millions of proteins from some sequence distribution, then performing an experiment that selects out the fittest sequences.
Sequence distributions commonly used today have relatively high entropy, and the resulting sequence diversity can lead to successful outcomes for a myriad of downstream selection experiments \cite{Bryant2021-yj, zhu2021aav}.
However, most of these sequences fail to assemble into a capsid that packages the genetic payload \cite{Adachi2014-if, Tse2017-ga, Ogden2019-eo}---a function called \emph{packaging}, which is the minimum requirement of a gene therapy delivery mechanism, and therefore a prerequisite to any other desiderata.

If sequence distributions could be developed with higher packaging rate, without compromising sequence diversity, then the success rate of downstream selection experiments should improve.
To this end, Zhu \& Brookes et al. \cite{zhu2021aav} use neural networks trained on sequence-packaging data to specify the parameters of sequence distributions that simultaneously have high entropy and yield sequences with high predicted packaging ability.
The sequences in this data varied at seven promising contiguous positions identified in previous work \cite{Dalkara2013-wy}, and elsewhere matched a wild type.
To accommodate commonly used DNA synthesis protocols, the authors parameterized their sequence distributions as independent categorical distributions over the four nucleotides at each of twenty-one contiguous sites, corresponding to codons at each of the seven sites of interest.

\subsubsection{Protocol for design experiments}

We followed the methodology of Zhu \& Brookes et al. \cite{zhu2021aav} to find sequence distributions with high mean predicted fitness---in particular, higher than that of the commonly used ``NNK'' sequence distribution \cite{Dalkara2013-wy}.
Specifically, we used their high-throughput data, which sampled millions of sequences from the NNK distribution and labeled each with an enrichment score quantifying its packaging fitness, based on its count before and after a packaging-based selection experiment. 
We introduced additional simulated measurement noise to these labels, where the parameters of the noise distribution were also estimated from the pre- and post-selection counts, resulting in labels ranging from $-7.53$ to $8.80$ for $8,552,729$ sequences (see \ifpnas S4 in the SI Appendix \else Section \ref{sect:app-aav} \fi for details).

We then randomly selected and held out one million of these data points, for calibration and test purposes described shortly, then trained a neural network on the remaining data to predict fitness from sequence.
Finally, following \cite{zhu2021aav}, we approximately solved an optimization problem that leveraged this regression model in order to specify the parameters of sequence distributions with high mean predicted fitness.
Specifically, let $\{p_\phi: \phi \in \Phi\}$ denote the class of sequence distributions parameterized as independent categorical distributions over the four nucleotides at each of twenty-one contiguous sequence positions.
We set the parameters of the designed sequence distribution by using stochastic gradient descent to approximately solve the following problem:
\begin{align}
\label{eq:aav}
    \phi_\lambda = \argmin_{\phi \in \Phi} D_\text{KL}(p_\lambda^\star || p_\phi)
\end{align}
where $p_\lambda ^\star(X) \propto \exp(\lambda \cdot \mu(X))$, $\mu$ is the neural network fit to the training data, and $\lambda \geq 0$ is an inverse temperature hyperparameter.
After solving for $\phi_\lambda$ for a range of inverse temperature values, $\lambda \in \{1, 2, 3, 4, 5, 6, 7\}$, we sampled designed sequences from $p_{\phi_\lambda}$ as described below, then used a randomized data splitting approach to construct confidence sets that achieve exact coverage.

\paragraph{Sampling designed sequences.}

Unlike in Section \ref{sect:protein-design}, here we did not have a label for every sequence in the input space---that is, all sequences that vary at the seven positions of interest, and that elsewhere match a wild type.
As an alternative, we used rejection sampling to sample from $p_{\phi_\lambda}$.
Specifically, recall that we held out a million of the labeled sequences.
The input space was sampled uniformly and densely enough by the high-throughput data set that we treated $990,000$ of these held-out labeled sequences as samples from a proposal distribution (that is, the NNK distribution) and were able to perform rejection sampling to sample designed sequences from $p_{\phi_\lambda}$ for which we have labels.

\paragraph{Constructing confidence sets for designed sequences.}

Note that rejection sampling results in some random number, at most $990,000$, of designed sequences; in practice, this number ranged from single digits to several thousand for $\lambda = 7$ to $\lambda = 1$, respectively.
To account for this variability, for each value of the inverse temperature, we performed $T = 500$ trials of the following steps.
We randomly split the one million held-out labeled sequences into $990,000$ proposal distribution sequences and $10,000$ sequences to be used as calibration data.
We used the former to sample some number of designed sequences, then used the latter to construct randomized staircase confidence sets (\ifpnas Alg. S1 in the SI Appendix) \else Alg. \ref{alg:staircase})\fi for each of the designed sequences.
The results we report next concern properties of these sets averaged over all $T = 500$ trials.

\subsubsection{Results}
\label{sect:aav-results}

\paragraph{Effect of inverse temperature.}

\begin{figure*}
  \includegraphics[width=17.8cm]{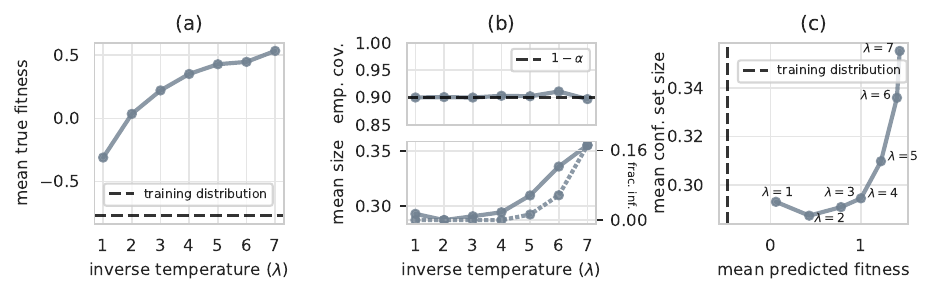}
  \caption{Quantifying uncertainty for predicted fitnesses of designed adeno-associated virus (AAV) capsid proteins. (a) Mean true fitness of designed sequences resulting from different values of the inverse temperature, $\lambda \in \{1, 2, \ldots, 7\}$. The dashed black line is the mean true fitness of sequences drawn from the NNK sequence distribution (i.e., the training distribution). (b) Top: empirical coverage of randomized staircase confidence sets (\ifpnas S1.D in the SI Appendix)\else Section \ref{app:rand-split}) \fi constructed for designed sequences. The dashed black line is the expected empirical coverage of $1 - \alpha = 0.9$. Bottom: fraction of confidence sets with infinite size (dashed gray line) and mean size of non-infinite confidence sets (solid gray line). The set size is reported as a fraction of the range of fitnesses in all the labeled data, $[-7.53, 8.80]$. (c) Trade-off between mean predicted fitness and mean confidence set size for $\lambda \in \{1, 2, \ldots, 7\}$. The dashed black line is the mean predicted fitness for sequences from the training distribution.}
  \label{fig:aav}
\end{figure*}

The inverse temperature hyperparameter, $\lambda$, in \eqref{eq:aav} plays a similar role as in Section \ref{sect:protein-design}: larger values result in designed sequences with higher mean true fitness (Fig.~\ref{fig:aav}a).
Note that the mean true fitness for all considered values of the inverse temperature is higher than that of the training distribution (the dashed black line, Fig.~\ref{fig:aav}a).

\paragraph{Empirical coverage and confidence set sizes.}

For all considered values of the inverse temperature, the empirical coverage of the confidence sets is very close to the expected value of $1 - \alpha = 0.9$ (Fig.~\ref{fig:aav}b, top).
Note that some designed sequences, which the neural network is particularly uncertain about, are given a confidence set with infinite size (Fig.~\ref{fig:aav}b, bottom).
The fraction of sets with infinite size, as well as the mean size of non-infinite sets, both increase with the inverse temperature (Fig.~\ref{fig:aav}b, bottom), which is consistent with our intuition that the neural network should be less confident about predictions that are much higher than fitnesses seen in the training data.

\paragraph{Using uncertainty quantification to set design procedure hyperparameters.}

As in Section \ref{sect:fluorescence-results}, the confidence sets we construct expose a trade-off between predicted fitness and predictive uncertainty as we vary the inverse temperature.
Generally, the higher the mean predicted fitness of the sequence distributions, the greater the mean confidence set size as well (Fig.~\ref{fig:aav}c).\footnote{The exception is the sequence distribution corresponding to $\lambda = 2$, which has a higher mean predicted fitness but on average smaller sets than $\lambda = 1$.
One likely explanation is that experimental measurement noise is particularly high for very low fitnesses, making low-fitness sequences inherently difficult to predict.}
One can inspect this trade-off to decide on an acceptable setting of the inverse temperature.
For example, observe that the mean set size does not grow appreciably between $\lambda = 1$ and $\lambda = 4$, even though the mean predicted fitness monotonically increases (Fig. \ref{fig:aav}b, bottom, \ref{fig:aav}c); similarly, the fraction of sets with infinite size also remains near zero for these values of $\lambda$ (Fig. \ref{fig:aav}b, bottom).
However, both of these quantities start to increase for $\lambda \geq 5$.
By $\lambda = 7$, for instance, more than $17\%$ of designed sequences are given a confidence set with infinite size, suggesting that $p_{\phi_7}$ has shifted too far from the training distribution for the neural network to be reasonably certain about its predictions.
Therefore, one might conclude that using $\lambda \in \{4, 5\}$ achieves an acceptable balance of designed sequences with higher predicted fitness than the training sequences and low enough predictive uncertainty.

% fmax = 8.798749497001769
% fmin = -7.530085215864544

\section{Discussion}

The predictions made by machine learning models are increasingly being used to make consequential decisions, which in turn influence the data that the models encounter.
Our work presents a methodology that allows practitioners to trust the predictions of learned models in such settings.
In particular, our protein design examples demonstrate how our approach can be used to navigate the trade-off between desirable predictions and predictive certainty inherent to design problems.

Looking beyond the design problem, the formalism of feedback covariate shift (FCS) introduced here captures a range of problem settings pertinent to modern-day deployments of machine learning.
In particular, FCS often occurs at each iteration of a feedback loop---for example, at each iteration of active learning, adaptive experimental design, and Bayesian optimization methods.
Applications and extensions of our approach to such settings are exciting directions for future investigation.

\ifpnas
\else
\section{Acknowledgments}

We are grateful to Danqing Zhu and David Schaffer for generously allowing us to work with their AAV packaging data, and to David H. Brookes and Akosua Busia for guidance on its analysis.
We also thank David H. Brookes, Hunter Nisonoff, Tijana Zrnic, and Meena Jagadeesan for insightful discussions and feedback. 
C.F. was supported by a National Science Foundation Graduate Research Fellowship under Grant DGE 2146752.
A.N.A. was partially supported by an National Science Foundation Graduate Research Fellowship and a Berkeley Fellowship.
S.B. was supported by the Foundations of Data Science Institute and the Simons Institute.
\fi

%% file: appendix.tex
\section{Proofs}

\subsection{Proof of Theorem \ref{thm}}
\label{app:proof}

Data from feedback covariate shift (FCS) are a special case of what we call \textit{pseudo-exchangeable}\footnote{The name \textit{pseudo-exchangeable} hearkens to the similarity of the factorized form to the pseudo-likelihood approximation of a joint density. Note, however, that each factor, $g_i(v_i; \, v_{-i})$, can only depend on the values and not the ordering of the other variables, $v_1, \ldots, v_{i - 1}, v_{i + 1}, \ldots, v_n$, whereas each factor in the pseudo-likelihood approximation also depends on the identities (i.e., the ordering) of the other variables.} random variables.

\begin{definition}
Random variables $V_1, \ldots, V_{n + 1}$ are \textup{pseudo-exchangeable} with factor functions $g_1, \ldots, g_{n + 1}$ and core function $h$ if the density, $f$, of their joint distribution can be factorized as
\begin{equation*}
    f(v_1, \ldots, v_{n + 1}) = \prod_{i = 1}^{n + 1} g_i(v_i; \, v_{-i}) \cdot h(v_1, \ldots, v_{n + 1}),
\end{equation*}
where $v_{-i} = v_{1:(n + 1)} \setminus v_i$,\footnote{With some abuse of notation, we denote $z_{-i} = z_{1:(n + 1)} \setminus z_i$ whenever possible, as done here, but use $z_{-i} = z_{1:n} \setminus z_i$ whenever we need to append a candidate test point, as done in the main text and in Theorem \ref{thm:conf} below. In either case, we will clarify.} each $g_i(\cdot; \, v_{-i})$ is a function that depends on the multiset $v_{-i}$ (that is, on the values in $v_{-i}$ but not on their ordering), and $h$ is a function that does not depend on the ordering of its $n + 1$ inputs.
\end{definition}

The following lemma characterizes the distribution of the scores of pseudo-exchangeable random variables, which allows for a pseudo-exchangeable generalization of conformal prediction in Theorem \ref{thm:conf}.
We then show that data generated under FCS are pseudo-exchangeable, and a straightforward application of Theorem \ref{thm:conf} yields Theorem \ref{thm} as a corollary.
Our technical development here builds upon the work of Tibshirani et al. \cite{tibshirani2019conformal}, who generalized conformal prediction to handle ``weighted exchangeable'' random variables, including data under standard covariate shift.

The key insight is that if we condition on the values, but not the ordering, of the scores, we can exactly describe their distribution. The following proposition is a generalization of arguments found in the proof of Lemma 3 in \cite{tibshirani2019conformal}; the subsequent result in Lemma \ref{lemma:quantile} is a generalization of that lemma.

\begin{proposition}
\label{prop:cond-dist}
Let $Z_1, \ldots, Z_{n + 1}$ be pseudo-exchangeable random variables with a joint density function, $f$, that can be written with factor functions $g_1, \ldots, g_{n + 1}$ and core function $h$. Let $S$ be any score function and denote $S_i = S(Z_i, Z_{-i})$ where $Z_{-i} = Z_{1:(n + 1)} \setminus \{Z_i\}$ for $i = 1 , \ldots, n + 1$. Define
\begin{align}
\label{eq:generic-weights}
    w_i(z_1, \ldots, z_{n + 1}) \equiv \frac{\sum_{\sigma: \sigma(n + 1) = i} \prod_{j = 1}^{n + 1} g_j(z_{\sigma(j)}; z_{-\sigma(j)})}{\sum_\sigma \prod_{j = 1}^{n + 1} g_j(z_{\sigma(j)}; z_{-\sigma(j)})}, \quad i = 1, \ldots, n + 1,
\end{align}
where the summations are taken over permutations, $\sigma$, of the integers $1, \ldots, n + 1$. For values $z = (z_1, \ldots, z_{n + 1})$, let $s_i = S(z_i, z_{-i})$ and let $E_z$ be the event that $\{Z_1, \ldots, Z_{n + 1}\} = \{z_1, \ldots, z_{n + 1}\}$ (that is, the multiset of values taken on by $Z_1, \ldots, Z_{n + 1}$ equals the multiset of the values in $z$). Then 
\begin{align}
    S_{n + 1} \mid E_z \sim \sum_{i = 1}^{n + 1} w_i(z_1, \ldots, z_{n + 1}) \, \delta_{s_i}.
\end{align}
\begin{proof}
For simplicity, we treat the case where $S_1, \ldots, S_{n + 1}$ are distinct almost surely; the result also holds in the general case, but the notation that accommodates duplicate values is cumbersome. For $i = 1, \ldots, n + 1$,
\begin{align}
\label{eq:pvconde}
    \Prob(S_{n + 1} = s_i \mid E_z) = \Prob(Z_{n + 1} = z_i \mid E_z) & = \frac{\sum_{\sigma: \sigma(n + 1) = i} f(z_{\sigma(1)}, \ldots, z_{\sigma(n + 1)})}{\sum_\sigma f(z_{\sigma(1)}, \ldots, z_{\sigma(n + 1)})} \\ 
    & = \frac{\sum_{\sigma: \sigma(n + 1) = i} \prod_{j = 1}^{n + 1} g_j(z_{\sigma(j)}; z_{-\sigma(j)}) \cdot h(z_{\sigma(1)}, \ldots, z_{\sigma(n + 1)}) }{\sum_{\sigma} \prod_{j = 1}^{n + 1} g_j(z_{\sigma(j)}; z_{-\sigma(j)}) \cdot h(z_{\sigma(1)}, \ldots, z_{\sigma(n + 1)})} \\
    & = \frac{\sum_{\sigma: \sigma(n + 1) = i} \prod_{j = 1}^{n + 1} g_j(z_{\sigma(j)}; z_{-\sigma(j)}) \cdot h(z_1, \ldots, z_{n + 1}) }{\sum_{\sigma} \prod_{j = 1}^{n + 1} g_j(z_{\sigma(j)}; z_{-\sigma(j)}) \cdot h(z_1, \ldots, z_{n + 1})} \\
    & = \frac{\sum_{\sigma: \sigma(n + 1) = i} \prod_{j = 1}^{n + 1} g_j(z_{\sigma(j)}; z_{-\sigma(j)})  }{\sum_{\sigma} \prod_{j = 1}^{n + 1} g_j(z_{\sigma(j)}; z_{-\sigma(j)})} \\
    & = w_i(z_1, \ldots, z_{n + 1}).
\end{align}
\end{proof}
\end{proposition}

\begin{lemma}
\label{lemma:quantile}
Let $Z_1, \ldots, Z_{n + 1}$ be pseudo-exchangeable random variables with a joint density function, $f$, that can be written with factor functions $g_1, \ldots, g_{n + 1}$ and core function $h$. Let $S$ be any score function and denote $S_i = S(Z_i, Z_{-i})$ where $Z_{-i} = Z_{1:(n + 1)} \setminus \{Z_i\}$ for $i = 1 , \ldots, n + 1$. For any $\beta \in (0, 1)$,
\begin{align}
    \Prob\left\{S_{n + 1} \leq \textsc{Quantile}_\beta \left( \sum_{i = 1}^{n + 1} w_i(Z_1, \ldots, Z_{n + 1})\, \delta_{S_i} \right) \right\} \geq \beta,
\end{align}
where $w_i(z_1, \ldots, z_{n + 1})$ is defined in \eqref{eq:generic-weights}.

\begin{proof}
Assume for simplicity of notation that $S_1, \ldots, S_{n + 1}$ are distinct almost surely (but the result holds generally). For data point values $z = (z_1, \ldots, z_{n + 1})$, let $s_i = S(z_i, z_{-i})$ and let $E_z$ be the event that $\{Z_1, \ldots, Z_{n + 1}\} = \{z_1, \ldots, z_{n + 1}\}$. By Proposition \ref{prop:cond-dist},
\begin{align}
    S_{n + 1} \mid E_z \sim \sum_{i = 1}^{n + 1} w_i(z_1, \ldots, z_{n + 1}) \, \delta_{s_i},
\end{align}
and consequently
\begin{align}
    \Prob \left(S_{n + 1} \leq \textsc{Quantile}_\beta \left( \sum_{i = 1}^{n + 1} w_i(z_1, \ldots, z_{n + 1}) \, \delta_{s_i}\right) \,\middle\vert\, E_z \right) \geq \beta,
\end{align}
by definition of the $\beta$-quantile; equivalently, since we condition on $E_z$,
\begin{align}
    \Prob \left(S_{n + 1} \leq \textsc{Quantile}_\beta \left( \sum_{i = 1}^{n + 1} w_i(Z_1, \ldots, Z_{n + 1}) \, \delta_{S_i}\right) \,\middle\vert\, E_z\right) \geq \beta.
\end{align}
Since this inequality holds for all events $E_z$, where $z$ is a vector of $n + 1$ data point values, smoothing gives
\begin{align}
    \Prob \left(S_{n + 1} \leq \textsc{Quantile}_\beta \left( \sum_{i = 1}^{n + 1} w_i(Z_1, \ldots, Z_{n + 1}) \, \delta_{S_i}\right) \right) \geq \beta.
\end{align}
\end{proof}
\end{lemma}

Lemma \ref{lemma:quantile} yields the following theorem, which enables a generalization of conformal prediction to pseudo-exchangeable random variables.

\begin{theorem}
\label{thm:conf}
Suppose $Z_1, \ldots, Z_{n + 1}$ where $Z_i = (X_i, Y_i) \in \mathcal{X} \times \reals$ are pseudo-exchangeable random variables with factor functions $g_1, \ldots, g_{n + 1}$. For any score function, $S$, and any miscoverage level, $\alpha  \in (0, 1)$, define for any point $x \in \mathcal{X}$:
\begin{align}
\label{eq:confset-generic}
    C_\alpha(x) & = \left \{ y \in \reals : S_{n + 1}(x, y) \leq \textsc{Quantile}_{1 - \alpha}\left(\sum_{i= 1}^{n + 1} w_i(Z_1, \ldots, Z_n, (x, y)) \, \delta_{S_i(x, y)} \right) \right \},
\end{align}
where $S_i(x, y) = S(Z_i, Z_{-i} \cup \{(x, y)\})$ and $Z_{-i} = Z_{1:n} \setminus Z_i$ for $i = 1, \ldots, n$, $S_{n + 1}(x, y) = S((x, y), Z_{1:n})$, and the weight functions $w_i$ are as defined in \eqref{eq:generic-weights}. Then $C_\alpha$ satisfies
\begin{align}
    \Prob\left( Y_{n + 1} \in C_\alpha(X_{n +1})\right) \geq 1 - \alpha,
\end{align}
where the probability is over all $n + 1$ data points, $Z_1, \ldots, Z_{n + 1}$. 

\begin{proof}
By construction, we have
\begin{align}
    Y_{n + 1} \in C_\alpha(X_{n + 1}) \iff S_{n + 1}(X_{n + 1}, Y_{n + 1}) \leq \textsc{Quantile}_{1 - \alpha}\left(\sum_{i = 1}^{n + 1} w_i(Z_1, \ldots, Z_{n + 1}) \, \delta_{S_i(X_{n + 1}, Y_{n + 1})}\right).
\end{align}
Applying Lemma \ref{lemma:quantile} gives the result.
\end{proof}
\end{theorem}

\sloppy
Finally, Theorem \ref{thm} follows as a corollary of Theorem \ref{thm:conf}. Denoting $Z_{n + 1} = Z_\text{test}$ and $Z_{-i} = Z_{1:(n +1)} \setminus \{Z_{i}\}$, observe that data, $(Z_1, \ldots, Z_{n + 1})$, under FCS are pseudo-exchangeable with the core function
\begin{align}
    h(z_1, \ldots, z_{n + 1}) & = \prod_{i = 1}^{n + 1} p_X(x_i) \, p_{Y \mid X}(y_i \mid x_i),
\end{align}
and factor functions $g_i(z_i; \, z_{-i}) = 1$ for $i = 1, \ldots, n$ and
\begin{align}
    g_{n + 1}(z_{n + 1}; \,z_{1:n}) & = \frac{\tilde{p}_{X; z_{1:n}}(x_{n + 1}) \, p_{Y \mid X}(y_{n + 1} \mid x_{n + 1})}{p_X(x_{n + 1}) \, p_{Y \mid X}(y_{n + 1} \mid x_{n + 1})} = \frac{\tilde{p}_{X; z_{1:n}}(x_{n + 1})}{p_X(x_{n + 1})} = v(x_{n + 1}; z_{1:n})
\end{align}
where $v(\cdot; \cdot)$ is the likelihood ratio function defined in \eqref{eq:ratio}.
The weights, $w_i(z_1, \ldots, z_{n + 1})$, in \eqref{eq:generic-weights} then simplify as
\begin{align}
    w_i(z_1, \ldots, z_{n + 1}) = \frac{\sum_{\sigma: \sigma(n + 1) = i} \prod_{j = 1}^{n + 1} g_j(z_{\sigma(j)}; z_{-\sigma(j)})}{\sum_\sigma \prod_{j = 1}^{n + 1} g_j(z_{\sigma(j)}; z_{-\sigma(j)})} & = \frac{\sum_{\sigma: \sigma(n + 1) = i} \prod_{j = 1}^{n + 1} g_j(z_{\sigma(j)}; z_{-\sigma(j)})}{\sum_{k = 1}^{n + 1} \sum_{\sigma: \sigma(n + 1) = k} \prod_{j = 1}^{n + 1} g_j(z_{\sigma(j)}; z_{-\sigma(j)})} \\
    & = \frac{\sum_{\sigma: \sigma(n + 1) = i} g_{n + 1}(z_{\sigma(n + 1)}; z_{-\sigma(n + 1)})}{\sum_{k = 1}^{n + 1} \sum_{\sigma: \sigma(n + 1) = k} g_{n + 1}(z_{\sigma(n + 1)}; z_{-\sigma(n + 1)})} \\
    & = \frac{\sum_{\sigma: \sigma(n + 1) = i} g_{n + 1}(z_i; z_{-i})}{\sum_{k = 1}^{n + 1} \sum_{\sigma: \sigma(n + 1) = k} g_{n + 1}(z_k; z_{-k})} \\
    & = \frac{n! \cdot g_{n + 1}(z_i; z_{-i})}{\sum_{k = 1}^{n + 1} n! \cdot g_{n + 1}(z_k; z_{-k})} \\
    & = \frac{v(x_i; z_{-i})}{\sum_{k = 1}^{n + 1} v(x_k; z_{-k})}.
\end{align}
These quantities are exactly the weight functions, $w_i^y$, defined in \eqref{eq:weights} and used in the full conformal confidence set in \eqref{eq:confset}: ${w_i^y(X_\text{test}) = w_i(Z_1, \ldots, Z_n, (X_\text{test}, y))}$ for $i = 1, \ldots, n + 1$.  That is, \eqref{eq:confset} gives the confidence set defined in \eqref{eq:confset-generic} for data under FCS. Applying Theorem \ref{thm:conf} then yields Theorem \ref{thm}.

\subsection{A randomized confidence set achieves exact coverage}
\label{app:random}

\sloppy
Here, we introduce the \textit{randomized $\beta$-quantile} and a corresponding randomized confidence set that achieves exact coverage.
To lighten notation, for a discrete distribution with probability masses $w = (w_1, \ldots, w_{n + 1})$ on points $s = (s_1, \ldots, s_{n + 1})$, where $s_i \in \reals$ and $w_i \geq 0, \sum_{i = 1}^{n + 1} w_i = 1$, we will write ${\textsc{Quantile}_\beta(s, w) = \textsc{Quantile}_\beta(\sum_{i = 1}^n w_i \delta_{s_i})}$.
Observe that $\textsc{Quantile}_\beta(s, w)$ is always one of the support points, $s_i$.
Now define the \textit{$\beta$-quantile lower bound}:
\begin{align}
\label{eq:lb}
    \textsc{QuantileLB}_\beta \left( s, w \right) & = \inf\left\{s: \sum_{i: s_i \leq s} w_i < \beta, \sum_{i: s_i \leq s} w_i + \sum_{j: s_j = \textsc{Quantile}_\beta(s, w)} w_j \geq \beta \right\},
\end{align}
which is either a support point strictly less than the $\beta$-quantile, or negative infinity.
Finally, letting $\textsc{QF}_\beta(s, w)$ and $\textsc{LF}_\beta(s, w)$ denote the CDF of the discrete distribution at $\textsc{Quantile}_\beta(s, w)$ and $\textsc{QuantileLB}_\beta(s, w))$, respectively, the randomized $\beta$-quantile is a random variable that takes on the value of either the $\beta$-quantile or the $\beta$-quantile lower bound:
\begin{align}
\label{eq:rq}
    \textsc{RandomizedQuantile}_\beta(s, w) & = \begin{cases}
       \textsc{QuantileLB}_\beta(s, w) & \text{w. p. $\frac{\textsc{QF}_\beta(s, w) - \beta}{\textsc{QF}_\beta(s, w) - \textsc{LF}_\beta(s, w)}$}, \\
       \textsc{Quantile}_\beta(s, w) & \text{w. p. $1 - \frac{\textsc{QF}_\beta(s, w) - \beta}{\textsc{QF}_\beta(s, w) - \textsc{LF}_\beta(s, w)}$}.
    \end{cases}
\end{align}

\sloppy
We use this quantity to define the \emph{randomized full conformal} confidence set, which, for any miscoverage level, $\alpha \in (0, 1)$, and $x \in \mathcal{X}$ is the following random variable:
\begin{align}
\label{eq:random-confset}
    C^\text{rand}_{\alpha}(x) & = \left \{y \in \reals : S((x, y), Z_{1:n}) \leq \textsc{RandomizedQuantile}_{1 - \alpha}(s(Z_1, \ldots, Z_n, (x, y)), w(Z_1, \ldots, Z_n, (x, y)) \right\},
\end{align}
where $s(Z_1, \ldots, Z_n, (x, y)) = (S_1, \ldots, S_n, S((x, y), Z_{1:n})$ and $S_i = S(Z_i, Z_{-i} \cup \{(x, y)\})$ for $i = 1, \ldots, n$, and ${w(Z_1, \ldots, Z_n, (x, y)) = (w_1^y(x), \ldots, w_{n + 1}^y(x))}$ where $w_i^y(x)$ is defined in \eqref{eq:weights}.
Note that for each candidate label, $y \in \reals$, an independent randomized $\beta$-quantile is instantiated; some values will use the $\beta$-quantile as the threshold on the score, while the others will use the $\beta$-quantile lower bound.
Randomizing the confidence set in this way yields the following result.
\begin{theorem}
\label{thm:random}
Suppose data, $Z_1, \ldots, Z_n, Z_\textup{test}$, are generated under feedback covariate shift and assume $\tilde{P}_{X; D}$ is absolutely continuous with respect to $P_X$ for all possible values of $D$. Then, for any miscoverage level, $\alpha \in (0, 1)$,  the randomized full confidence set, $C_\alpha^\textup{rand}$, in \eqref{eq:random-confset} satisfies the \textup{exact coverage} property:
\begin{align}
\label{eq:exact-coverage}
    \Prob(Y_\textup{test} \in C_\alpha^\textup{rand}(X_\textup{test})) = 1 - \alpha,
\end{align}
where the probability is over $Z_1, \ldots, Z_n, Z_\textup{test}$ and the randomness in $C_\alpha^\textup{rand}$.

\begin{proof} Denote $Z_{n + 1} = Z_\text{test}$ and $Z = (Z_1, \ldots, Z_{n + 1})$. For a vector of $n + 1$ data point values, $z = (z_1, \ldots, z_{n + 1})$, use the following shorthand:
\begin{align*}
  & Q_\beta(z) = \textsc{Quantile}_\beta(s(z), w(z)), \\
  & L_\beta(z) = \textsc{QuantileLB}_\beta(s(z), w(z)), \\
  & R_\beta(z) = \textsc{RandomizedQuantile}_\beta(s(z), w(z)), \\
  & \textsc{QF}_\beta(z) = \textsc{QF}_\beta(s(z), w(z)), \\
  & \textsc{LF}_\beta(z) = \textsc{LF}_\beta(s(z), w(z)).
\end{align*}
As in the proof of Lemma \ref{lemma:quantile}, consider the event, $E_z$, that $\{Z_1, \ldots, Z_{n + 1}\} = \{z_1, \ldots, z_{n + 1}\}$. Assuming for simplicity that the scores are distinct almost surely, by Proposition \ref{prop:cond-dist}
\begin{align}
    S(Z_{n + 1}, Z_{1 : n}) \mid E_z \sim \sum_{i = 1}^{n + 1} w_i(z_1, \ldots, z_{n + 1}) \, \delta_{S(z_i, z_{-i})},
\end{align}
and consequently
\begin{align}
    & \Prob( S(Z_{n + 1}, Z_{1 : n}) \leq \textsc{R}_{1 - \alpha}(z) \mid E_z) \\
    & = \Prob( S(Z_{n + 1}, Z_{1 : n}) \leq \textsc{R}_{1 - \alpha}(z) \mid E_z, \textsc{R}_{1 - \alpha}(z) = \textsc{Q}_{1 - \alpha}(z)) \cdot  \Prob(\textsc{R}_{1 - \alpha}(z) = \textsc{Q}_{1 - \alpha}(z) \mid E_z) + \\
    & \qquad \Prob( S(Z_{n + 1}, Z_{1 : n}) \leq \textsc{R}_{1 - \alpha}(z) \mid E_z, \textsc{R}_{1 - \alpha}(z) = \textsc{L}_{1 - \alpha}(z)) \cdot  \Prob(\textsc{R}_{1 - \alpha}(z) = \textsc{L}_{1 - \alpha}(z) \mid E_z) \\
    & = \Prob( S(Z_{n + 1}, Z_{1 : n}) \leq \textsc{Q}_{1 - \alpha}(z) \mid E_z) \cdot \left(1 - \frac{\textsc{QF}_{1 - \alpha}(z) - (1 - \alpha)}{\textsc{QF}_{1 - \alpha}(z) - \textsc{LF}_{1 - \alpha}(z)}\right) + \\
    & \qquad \Prob( S(Z_{n + 1}, Z_{1 : n}) \leq \textsc{L}_{1 - \alpha}(z) \mid E_z) \cdot \frac{\textsc{QF}_{1 - \alpha}(z) - (1 - \alpha)}{\textsc{QF}_{1 - \alpha}(z) - \textsc{LF}_{1 - \alpha}(z)} \\
    & = \textsc{QF}_{1 - \alpha}(z) \cdot \left( 1 - \frac{\textsc{QF}_{1 - \alpha}(z) - (1 - \alpha)}{\textsc{QF}_{1 - \alpha}(z) - \textsc{LF}_{1 - \alpha}(z)}\right) + \textsc{LF}_{1 - \alpha}(z) \cdot  \frac{\textsc{QF}_{1 - \alpha}(z) - (1 - \alpha)}{\textsc{QF}_{1 - \alpha}(z) - \textsc{LF}_{1 - \alpha}(z)} \\
    & = -\left( \textsc{QF}_{1 - \alpha}(z) - \textsc{LF}_{1 - \alpha}(z)\right) \cdot \frac{\textsc{QF}_{1 - \alpha}(z) - (1 - \alpha)}{\textsc{QF}_{1 - \alpha}(z) - \textsc{LF}_{1 - \alpha}(z)} + \textsc{QF}_{1 - \alpha}(z) \\
    & = - \textsc{QF}_{1 - \alpha}(z) + (1 - \alpha) + \textsc{QF}_{1 - \alpha}(z) \\
    & = 1 - \alpha.
\end{align}
Since we condition on $E_z$, we equivalently have
\begin{align}
    \Prob(S(Z_{n + 1}, Z_{1:n}) \leq R_{1 - \alpha}(Z) \mid E_z) & = 1 - \alpha,
\end{align}
and since this equality holds for all events $E_z$, where $z$ is a vector of $n + 1$ data point values, taking an expectation over $E_z$ yields
\begin{align}
    \Prob(S(Z_{n + 1}, Z_{1:n}) \leq R_{1 - \alpha}(Z)) & = 1 - \alpha.
\end{align}
Finally, since
\begin{align}
    Y_{n + 1} \in C_\alpha^\textup{rand}(X_{n + 1}) \iff S(Z_{n + 1}, Z_{1:n}) \leq R_{1 - \alpha}(Z),
\end{align}
the result follows.
\end{proof}
\end{theorem}
Note that standard covariate shift is subsumed by feedback covariate shift, so Theorem \ref{thm:random} can be used to construct a randomized confidence set with exact coverage under standard covariate shift as well.

\subsection{Data splitting}
\label{app:split}

In general, computing the full conformal confidence set, $C_\alpha(x)$, using Alg. \ref{alg:generic} requires fitting $(n + 1) \times |\mathcal{Y}|$ regression models.
A much more computationally attractive alternative is called a \emph{data splitting} or \emph{split conformal} approach~\cite{papadopoulos2002inductive,lei2018distribution}, in which we (i) randomly partition the labeled data into disjoint training and \emph{calibration} data sets, (ii) fit a regression model to the training data, and (iii) use the scores that it provides for the calibration data (but not the training data) to construct confidence sets for test data points.
Though this approach only requires fitting a single model, the trade-off is that it does not use the labeled data as efficiently: only some fraction of our labeled data can be used to train the regression model.
This limitation may be inconsequential for settings with abundant data, but can be a nonstarter when labeled data is limited, such as in many protein design problems.

Here, we show how data splitting simplifies feedback covariate shift (FCS) to standard covariate shift.
We then use the data splitting method from Tibshirani et al. \cite{tibshirani2019conformal} to produce confidence sets with coverage; the subsequent subsection shows how to introduce randomization to achieve exact coverage.

To begin, we recall the standard covariate shift model \cite{shimodaira2000, Sugiyama2005input, sugiyama2007covariate}.
The training data, $Z_1, \ldots, Z_n$ where $Z_i = (X_i, Y_i)$, are i.i.d. from some distribution: $X_i \sim P_X, Y_i \sim P_{Y \mid X_i}$ for $i = 1, \ldots, n$.
A test data point, $Z_\text{test} = (X_\text{test}, Y_\text{test})$, is drawn from a different input distribution but the same conditional distribution, $X_\text{test} \sim \tilde{P}_X, Y_\text{test} \sim P_{Y \mid X_\textup{test}}$, independently from the training data.
In contrast to FCS, here the test input cannot be chosen in a way that depends on the training data.

Returning to FCS, suppose we randomly partition all our labeled data into disjoint training and calibration data sets.
Let $\mu$ denote the regression model fit to the training data; we henceforth consider $\mu$ as fixed and make no further use of the training data.
As such, without loss of generality we will use $Z_1, \ldots, Z_{m}$ to refer to the calibration data.
Now suppose the test input distribution is induced by the trained regression model, $\mu$; we write this as $\tilde{P}_{X; \mu}$.
Observe that, conditioned on the training data, we now have a setting where the calibration and test data are drawn from different input distributions but the same conditional distribution, $P_{Y \mid X}$, and are independent of each other. That is, data splitting returns us to standard covariate shift.

To construct valid confidence sets under standard covariate shift, define the following likelihood ratio function:
\begin{align}
\label{eq:scs-lr}
    v(x) & = \frac{\tilde{p}_{X; \mu}(x)}{p_X(x)},
\end{align}
where $p_X$ and $\tilde{p}_{X; \mu}$ refer to the densities of the training and test input distributions, respectively.
We restrict our attention to score functions of the following form \cite{angelopoulos2021gentle}:
\begin{align}
\label{eq:score}
    S(x, y) & = \frac{|y - \mu(x)|}{u(x)}.
\end{align}
where $u$ is any heuristic, nonnegative notion of uncertainty; one can also set $u(x) = 1$ to recover the residual score function. 
Note that, since we condition on the training data and treat the regression model as fixed, the score of a point, $(x, y)$, is no longer also a function of other data points.
Finally, for any miscoverage level, $\alpha \in (0, 1)$, and any $x \in \mathcal{X}$, define the \emph{split conformal} confidence set as
\begin{align}
\begin{split}
    \label{eq:split-confset}
    & C^\text{split}_\alpha(x) = \mu(x) \pm q \cdot u(x), \\
    & q = \textsc{Quantile}_{1 - \alpha}\left(\sum_{i = 1}^{m} w_i(x) \, \delta_{S_i} + w_{n + 1}(x) \, \delta_\infty \right),
\end{split}
\end{align}
where $S_i = S(X_i, Y_i)$ for $i = 1, \ldots, m$ and
\begin{align}
\label{eq:scs-weight}
    w_i(x) & = \frac{v(X_i)}{\sum_{j = 1}^{m} v(X_j) + v(x)}, \quad i = 1, \ldots, {m}, \\
    w_{m + 1}(x) & = \frac{v(x)}{\sum_{j=1}^{m} v(X_j) + v(x)}.
\end{align}
For data under standard covariate shift, the split conformal confidence set achieves coverage, as first shown in \cite{tibshirani2019conformal}.
\begin{theorem}[Corollary 1 in \cite{tibshirani2019conformal}]
\label{thm:split}
Suppose calibration and test data, $Z_1, \ldots, Z_m, Z_\textup{test}$, are under standard covariate shift, and assume $\tilde{P}_{X;\mu}$ is absolutely continuous with respect to $P_X$. For score functions of the form in \eqref{eq:score}, and any miscoverage level, $\alpha \in (0, 1)$, the split conformal confidence set, $C^\textup{split}_\alpha(x)$, in \eqref{eq:split-confset} satisfies the coverage property in \eqref{eq:coverage}.
\end{theorem}
To achieve exact coverage, we can introduce randomization, as we discuss next.

\subsection{Data splitting with randomization achieves exact coverage}
\label{app:rand-split}

Here, we stay in the setting and notation of the previous subsection and demonstrate how randomizing the $\beta$-quantile enables a data splitting approach to achieve exact coverage.
For any score function of the form in \eqref{eq:score}, any miscoverage level, $\alpha \in (0, 1)$, the \emph{randomized split conformal} confidence set is the following random variable for $x \in \mathcal{X}$:
\begin{align}
\label{eq:rand-split-confset}
 C^{\textup{rand,split}}_{\alpha}(x) = \left \{ y \in \reals : S(x, y) \leq \textsc{RandomizedQuantile}_{1 - \alpha} \left((S_1, \ldots, S_{m}, S(x, y)), (w_1(x), \ldots, w_{m + 1}(x)) \right) \right\},
\end{align}
where the randomized $\beta$-quantile, $\textsc{RandomizedQuantile}_\beta$ is defined in \eqref{eq:rq}, $S_i = S(X_i, Y_i)$ for $i = 1, \ldots, {m}$, and $w_i(\cdot)$ for $i = 1, \ldots, m + 1$ is defined in \eqref{eq:scs-weight}. Observe that for each candidate label, $y \in \reals$, an independent randomized $\beta$-quantile is drawn, such that the scores of some values are compared to the $\beta$-quantile while the others are compared to the $\beta$-quantile lower bound.
The exact coverage property of this confidence set is a consequence of Theorem \ref{thm:random}.

\begin{corollary}
\label{cor:rand-split}
Suppose calibration and test data, $Z_1, \ldots, Z_m, Z_\textup{test}$, are under standard covariate shift, and assume $\tilde{P}_{X; \mu}$ is absolutely continuous with respect to $P_X$. For score functions of the form in \eqref{eq:score}, and any miscoverage level, $\alpha \in (0, 1)$, the randomized split conformal confidence set, $C^\textup{rand,split}_\alpha(x)$, in \eqref{eq:rand-split-confset} satisfies the exact coverage property in \eqref{eq:exact-coverage}.
\begin{proof}
Since standard covariate shift is a special case of FCS, the calibration and test data can be described by FCS where $\tilde{P}_{X; D} = \tilde{P}_{X; \mu}$ for any multiset $D$.
The randomized split conformal confidence set, $C^\textup{rand,split}_\alpha$, is simply the randomized full conformal confidence set, $C^\textup{rand}_\alpha$, defined in \eqref{eq:random-confset}, instantiated with the scores $S((x, y), Z_{1:m}) = S(x, y)$ and $S(Z_i, Z_{-i} \cup \{(x, y)\}) = S(Z_i)$ for $i = 1, \ldots, m$, and weights resulting from $\tilde{P}_{X; D} = \tilde{P}_{X; \mu}$ for all $D$. The result then follows from Theorem \ref{thm:random}.
\end{proof}
\end{corollary}

While we only need to fit a single regression model to compute the scores for data splitting, naively it might seem that in practice, we need to approximate $C^{\textup{rand,split}}_{\alpha}(x)$ by introducing a discrete grid of candidate labels, $\mathcal{Y} \subset \reals$, and computing a randomized $\beta$-quantile for $|\mathcal{Y}|$ different discrete distributions.
Fortunately, we can construct an alternative confidence set that also achieves exact coverage, the \emph{randomized staircase} confidence set, $C^{\textup{staircase}}_{\alpha}$, which only requires sorting $m$ scores and an additional $O(m)$ floating point operations to compute (see Alg. \ref{alg:staircase}).

\begin{figure}[hbt!]
  \centering
  \includegraphics[width=15cm]{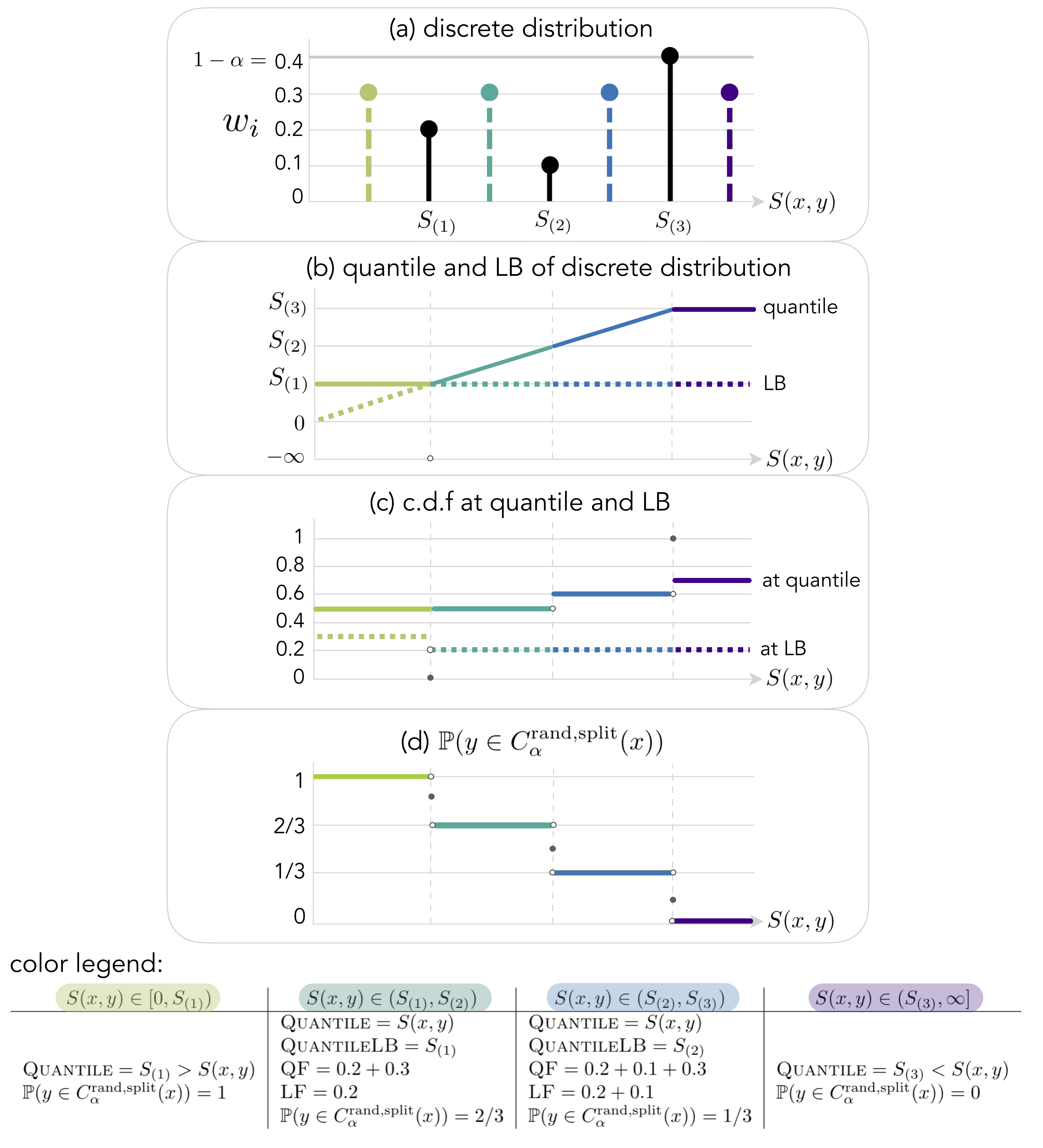}
  \caption{Depiction of how the probability $\Prob(y \in C_\alpha^\textup{rand,split}(x))$ is a piecewise constant function of $y$. (a) Given the values of the calibration data and test input, the scores $S_1, \ldots, S_m$ and corresponding probability masses $w_1, \ldots, w_m$ (black stems), as well as the probability mass for the test input, $w_{m + 1} = 0.3$, are fixed. The only quantity that depends on $y$ is $S(x, y)$. Four example values are shown as dashed green, teal, blue, and purple stems, representing values in $[0, S_{(1)}), (S_{(1)}, S_{(2)}), (S_{(2)}, S_{(3)})$, and $(S_{(3)}, \infty]$, respectively (see color legend). Note that in this example, $1 - \alpha = 0.4$. (b) The $0.4$-quantile and $0.4$-quantile lower bound of the discrete distribution in the top panel as a function of $S(x, y)$, where the colors correspond to values of $S(x,y)$ in the intervals just listed. Note the discontinuity in the $0.4$-quantile lower bound at $S(x, y) = S_{(1)}$. (c) The c.d.f. of the discrete distribution at the $0.4$-quantile and $0.4$-quantile lower bound. Note the discontinuities when $S(x, y)$ equals a calibration score. (d) The probability $\Prob(y \in C_\alpha^\textup{rand,split}(x))$, which equals $1$ or $0$ if $S(x, y) = 0.4$-quantile lower bound or $S(x, y) > 0.4$-quantile, respectively, and otherwise equals the probability in \eqref{eq:rq} that the randomized $0.4$-quantile equals the $0.4$-quantile: $1 - \frac{\textsc{QF} - 0.4}{\textsc{QF} - \textsc{LF}}$, where $\textsc{QF}$ and $\textsc{LF}$ denote the c.d.f. at the $0.4$-quantile and $0.4$-quantile lower bound, respectively. Color legend: calculations of the plotted quantities (calculations for $S(x, y) = S_{(i)}$ omitted).}
  \label{fig:staircase}
\end{figure}

\begin{algorithm}[hbt!]
\footnotesize
\caption{Randomized staircase confidence set}
\label{alg:staircase}

\textbf{Input:} Miscoverage level, $\alpha \in (0, 1)$; calibration data, $Z_1, \ldots, Z_m$, where $Z_i = (X_i, Y_i)$; test input, $X_\textup{test}$; subroutine for likelihood ratio function, $v(\cdot)$, defined in \eqref{eq:scs-lr};
subroutine for uncertainty heuristic, $u(\cdot)$; subroutine for regression model prediction, $\mu(\cdot)$. 

\textbf{Output:} Randomized staircase confidence set, $C = C^\textup{staircase}_\alpha(X_\textup{test})$.

\begin{algorithmic}[1]

\For{$i = 1, \ldots, m$} \Comment{Compute calibration scores}
\State $S_i \gets |Y_i - \mu(X_i)| / u(X_i)$
\State $v_i \gets v(X_i)$
\EndFor
\State $v_{m + 1} \gets v(X_\textup{test})$
\For{$i = 1, \ldots, m + 1$} \Comment{Compute calibration and test weights}
\State $w_i \gets v_i / \sum_{j = 1}^{m + 1} v_j$
\EndFor

\State $C \gets \emptyset$

\State $\texttt{LowerBoundIsSet} \gets \texttt{False}$

\State $S_{(0)} = 0, w_0 = 0$ \Comment{Dummy values so for-loop will include $[0, S_{(1)}]$}

\For{$i = 0, \ldots, m - 1$}

\If{$\sum_{j: S_j \leq S_{(i)}} w_j + w_{m + 1} < 1 - \alpha$} \Comment{$S(x, y) \leq \beta$-quantile lower bound, so include deterministically}

\State $C = C \cup \left[\mu(X_\textup{test}) +  S_{(i)} \cdot u(X_\textup{test}), \mu(X_\textup{test}) + S_{(i + 1)} \cdot u(X_\textup{test})\right] \cup \left[\mu(X_\textup{test}) -  S_{(i + 1)} \cdot u(X_\textup{test}), \mu (X_\textup{test}) - S_{(i)} \cdot u(X_\textup{test})\right]$ \label{ln:staircase1}

\ElsIf{$\sum_{j: S_j \leq S_{(i)}} w_j + w_{m + 1} \geq 1 - \alpha$ and $\sum_{j: S_j \leq S_{(i)}} w_j < 1 - \alpha$} \Comment{$S(x, y) = \beta$-quantile, so randomize inclusion}

\If{$\texttt{LowerBoundIsSet} = \texttt{False}$}
\State $\texttt{LowerBoundIsSet} \gets \texttt{True}$  \Comment{Set $\beta$-quantile lower bound}
\State $LF = \sum_{j: S_j \leq S_{(i)}} w_j$
\EndIf

\State $F \gets \frac{\sum_{j: S_j \leq S_{(i)}} w_j + w_{m + 1} - (1 - \alpha)}{\sum_{j: S_j \leq S_{(i)}} w_j + w_{m + 1} - LF}$ 
\State $b \sim \textup{Bernoulli}(1 - F)$

\If{$b$}
\State $C = C \cup \left[\mu(X_\textup{test}) +  S_{(i)} \cdot u(X_\textup{test}), \mu(X_\textup{test}) + S_{(i + 1)} \cdot u(X_\textup{test})\right] \cup \left[\mu(X_\textup{test}) -  S_{(i + 1)} \cdot u(X_\textup{test}), \mu(X_\textup{test}) - S_{(i)} \cdot u(X_\textup{test})\right]$ \label{ln:staircase2}
\EndIf

\EndIf

\EndFor 

\If{$\sum_{i = 1}^m w_i < 1 - \alpha$} \Comment{For $S(x, y) > S_{(m)}$, either $S(x, y) = \beta$-quantile or $S(x, y) > \beta$-quantile}

\If{$\texttt{LowerBoundIsSet} = \texttt{False}$}
\State $LF = \sum_{i = 1}^m w_i$
\EndIf

\State $F \gets \frac{1 - (1 - \alpha)}{1 - LF}$
\State $b \sim \textup{Bernoulli}(1 - F)$

\If{$b$}
\State $C = C \cup \left[\mu(X_\textup{test}) +  S_{(m)} \cdot u(X_\textup{test}), \infty \right] \cup \left[-\infty, \mu(X_\textup{test}) - S_{(m)} \cdot u(X_\textup{test})\right]$ \label{ln:staircase3}
\EndIf
\EndIf

\end{algorithmic}
\end{algorithm}

At a high level, its construction is based on the observation that for any $x \in \mathcal{X}$ and $y \in \reals$, the quantity $\Prob(y \in C_\alpha^\textup{rand,split}(x))$, where the probability is over the randomness in $C_\alpha^\textup{rand,split}(x)$, is a piecewise constant function of $y$.
Instead of testing each value of $y \in \reals$, we can then construct this piecewise constant function, and randomly include entire intervals of $y$ values that have the same value of $\Prob(y \in C_\alpha^\textup{rand,split}(x))$.

Fig. \ref{fig:staircase} illustrates this observation, which we now explain.
First, the discrete distribution in \eqref{eq:rand-split-confset} has probability masses $w_1(x), \ldots, w_{m + 1}(x)$ at the points $S_1, \ldots, S_m, S(x, y)$, respectively.
Given the values of the $m$ calibration data points and the test input, $x$, all of these quantities are fixed---except for the score of the candidate test data point, $S(x, y)$.
That is, the only quantity that depends on the value of $y$ is $S(x, y)$, which is the location of the probability mass $w_{m + 1}(x)$; the remaining $m$ support points and their corresponding probability masses do not not change with $y$.

Now consider the calibration scores, $S_1, \ldots, S_m$, sorted in ascending order.
Observe that for any pair of successive sorted scores, $S_{(i)}$ and $S_{(i + 1)}$, the entire interval of $y$ values such that $S(x, y) \in (S_{(i)}, S_{(i + 1})$ belongs to one of three categories: $S(x, y) \leq \beta$-quantile lower bound (of the discrete distribution with probability masses $w_1, \ldots, w_{m + 1}$ at support points $S_1, \ldots, S_m, S(x, y)$), $S(x, y) = \beta$-quantile, or $S(x, y) > \beta$-quantile.
An interval of $y$ values that belongs to the first category is deterministically included in $C_\alpha^\textup{rand,split}(x)$, regardless of the randomness in the randomized $\beta$-quantile (color-coded green in Fig. \ref{fig:staircase}), while an interval that belongs to the last category is deterministically excluded (color-coded purple in Fig. \ref{fig:staircase}).
The only $y$ values whose inclusion is not deterministic are those in the second category (color-coded teal and blue), which are randomly included with the probability, given in \eqref{eq:rq}, that the randomized $\beta$-quantile equals the $\beta$-quantile.
Consequently, we can identify the intervals of $y$ values belonging to each of these categories, and for those in the second category, compute the probability that the randomized $\beta$-quantile is instantiated as the $\beta$-quantile, which is $\Prob(y \in C_\alpha^\textup{rand,split}(x))$.

This probability turns out to be a piecewise constant function of $y$.
Note that it is computed from two quantities: the c.d.f. at the $\beta$-quantile and the c.d.f at the $\beta$-quantile lower bound (see \eqref{eq:rq}).
As depicted in Fig. \ref{fig:staircase} (third panel from top), for any two successive sorted calibration scores, $S_{(i)}$ and $S_{(i + 1)}$, both of these quantities are constant over $S(x, y) \in (S_{(i)}, S_{(i + 1})$.
That is, both the c.d.f. at the $\beta$-quantile and the c.d.f. at $\beta$-quantile lower bound are piecewise constant functions of $y$, which only change values at the calibration scores, $S_1, \ldots, S_m$ (and can take on different values exactly at the calibration scores).
Consequently, the probability $\Prob(y \in C_\alpha^\textup{rand,split}(x))$ is also a piecewise constant function of $y$, which only changes values at the calibration scores.
It attains its highest value at $\mu(x)$ and decreases as $y$ moves further away from it, resembling a staircase, as depicted in Fig. \ref{fig:staircase} (fourth panel from the top).

\begin{sloppypar}
Therefore, instead of computing a randomized $\beta$-quantile for all $y \in \reals$, we can simply compute the value of this probability on the $m + 1$ intervals between neighboring sorted calibration scores: $[0, S_{(1)}), (S_{(1)}, S_{(2)}), \ldots, (S_{(m - 1)}, S_{(m)}), (S_{(m)}, \infty]$, as well as its value exactly at the $m$ calibration scores.
These probabilities may equal $1$ or $0$, which correspond to the two cases earlier described wherein $y$ is deterministically included or excluded, respectively.
If the probability is not $1$ or $0$, then we can randomly include the entire set of values of $y$ such that $S(x, y)$ falls in the interval.
Due to the form of the score in \eqref{eq:score}, this set comprises two equal-length intervals on both sides of $\mu(x)$: {$(\mu(x) - S_{(i + 1)}, \mu(x) - S_{(i)}) \cup (\mu(x) + S_{(i + 1)}, \mu(x) + S_{(i)})$}.
\end{sloppypar}

Finally, if we assume that scores are distinct almost surely, then our treatment of values of $y$ such that $S(x, y) = S_i$ for $i = 1, \ldots, m$, does not affect the exact coverage property.
For simplicity, Alg. \ref{alg:staircase} therefore includes or excludes closed intervals that contain these $y$ values as endpoints, rather than treating them separately.

\paragraph{More general score functions.} In the reasoning above, we use the assumption that the score function takes the form in~\eqref{eq:score} only at the end of the argument, to infer the form of the sets of $y$ values.
We can relax this assumption as follows.
For any continuous score function, consider the preimage of the intervals $[0, S_{(1)}), (S_{(1)}, S_{(2)}), \ldots, (S_{(m - 1)}, S_{(m)}), (S_{(m)}, \infty]$ under the function $S(x, \cdot)$ (a function of the second argument with $x$ held fixed), rather than the intervals given explicitly in Lines \ref{ln:staircase1}, \ref{ln:staircase2}, and \ref{ln:staircase3} of Alg.~\ref{alg:staircase}.
This algorithm then gives exact coverage for any continuous score function, although it will only be computationally feasible when these preimages can be computed efficiently.

\section{Efficient computation of the full conformal confidence set for ridge regression and Gaussian process regression}
\label{app:efficient}

\subsection{Ridge regression}

When the likelihood of the test input is a function of the prediction from a ridge regression model, it is possible to compute the scores and weights for the full conformal confidence set by fitting $n + 1$ models and $O(n \cdot p \cdot |\mathcal{Y}|)$ additional floating point operations, instead of naively fitting $(n + 1) \times \mathcal{Y}$ models, as demonstrated in Alg. \ref{alg:ridge}.

For the fluorescent protein design experiments, the \textsc{TestInputLikelihood} subroutine in Alg. \ref{alg:ridge} computed the likelihood in \eqref{eq:protein-test-dist}, that is,
\begin{align}
\begin{split}
\label{eq:likelihood-computation}
    \textsc{TestInputLikelihood}(a_i + b_i y) & \gets \frac{\exp(\lambda \cdot (a_i + b_i y))}{\cdot \sum_{x \in \mathcal{X}} \exp(\lambda \cdot (C_i + y \mathbf{A}_{-i, n})^T x) }, \\
    \textsc{TestInputLikelihood}(a_{n + 1}) & \gets \frac{\exp(\lambda \cdot a_{n + 1})}{ \cdot \sum_{x \in \mathcal{X}} \exp(\lambda \cdot \beta^T x )},
\end{split}
\end{align}
where the input space $\mathcal{X}$ was the combinatorially complete set of $8,192$ sequences.
The $\textsc{TrainInputLikelihood}$ subroutine returned the likelihood under the training input distribution, which is simply equal to to $1 / 8192$, since training sequences were sampled uniformly from the combinatorially complete data set.
See \texttt{\href{https://github.com/clarafy/conformal-for-design}{https://github.com/clarafy/conformal-for-design}} for an implementation.

Computing the test input likelihoods was dominated by the $(n + 1) \times |\mathcal{Y}|$ normalizing constants, which can be computed efficiently using a single tensor product between an $(n + 1) \times p \times |\mathcal{Y}|$ tensor containing the model parameters, $C_i + y \mathbf{A}_{-i, n}$ and $\beta$, and an $|\mathcal{X}| \times p$ data matrix containing all inputs in $\mathcal{X}$. For domains, $\mathcal{X}$, that are too large for the normalizing constants to be computed exactly, one can turn to tractable Monte Carlo approximations.

\begin{algorithm}[hbt!]
\footnotesize
\caption{Efficient computation of scores and weights for ridge regression-based feedback covariate shift}\label{alg:ridge}

\textbf{Input:} training data, $Z_1, \ldots, Z_n$, where $Z_i = (X_i, Y_i)$; test input, $X_{n + 1}$; grid of candidate labels, $\mathcal{Y} \subset \reals$; subroutine for test input likelihood, $\textsc{TestInputLikelihood}(\cdot)$, that takes an input's predicted fitness and outputs its likelihood under the test input distribution; subroutine for training input likelihood, $\textsc{TrainInputLikelihood}(\cdot)$.

\textbf{Output:} scores $S_i(X_{n + 1}, y)$ and likelihood ratios $v(X_i, \augloodata{i})$ for $i = 1, \ldots, n + 1$, $y \in \mathcal{Y}$.

\begin{algorithmic}[1]

% \Comment{Train $n + 1$ models (compute the linear parameterizations of $\augloomu(X_i)$ as functions of $y$).}

\For{$i = 1, \ldots, n$}
\State $C_i \gets \sum_{j = 1}^{n - 1} Y_{-i; j} \mathbf{A}_{-i; j}$ 
\State $a_i \gets C_i^T X_i$
\State $b_i \gets \mathbf{A}_{-i; n}^T X_i$
\EndFor

\State $\beta \gets (\mathbf{X}^T\mathbf{X} + \gamma I)^{-1} \mathbf{X}^T Y$

\State $ a_{n + 1} \gets \beta^T X_{n + 1}$

% \Comment{For each $X_i$, compute scores and weights for all $y \in \mathcal{Y}$.}

\For{$i = 1, \ldots, n$} 
\For{$y \in \mathcal{Y}$}
\State $S_i(X_{n + 1}, y) \gets |Y_i - (a_i + b_i y)|$ \Comment{Can vectorize via outer product between $(b_1, \ldots, b_n)$ and vector of all $y \in \mathcal{Y}$.}

\State $ v(X_i; Z_{-i, y}) \gets \textsc{TestInputLikelihood}(a_i + b_i y) / \textsc{TrainInputLikelihood}(X_i)$ \Comment{Can vectorize (see commentary on \eqref{eq:likelihood-computation}).} \label{line:test-input1}
\EndFor
\EndFor

\State $S_{n + 1}(X_{n + 1}, y) \gets |y - a_{n + 1}|$

\State $ v(X_{n + 1}; Z_{1:n}) \gets \textsc{TestInputLikelihood}(a_{n + 1}) / \textsc{TrainInputLikelihood}(X_{n + 1})$ \label{line:test-input2}

\end{algorithmic}
\end{algorithm}

\subsection{Gaussian process regression}
\label{app:gpr}

Here we describe how the scores and weights for the confidence set in \eqref{eq:confset} can be computed efficiently, when the likelihood of the test input distribution is a function of the predictive mean and variance of a Gaussian process regression model.

For an arbitrary kernel and two data matrices, $\mathbf{V} \in \reals^{n_1 \times p}$ and $\mathbf{V'} \in \reals^{n_2 \times p}$, let $K(\mathbf{V}, \mathbf{V}')$ denote the $n_1 \times n_2$ matrix where the $(i, j)$-th entry is the covariance between the $i$-th row of $\mathbf{V}$ and $j$-th row of $\mathbf{V}'$.
The mean prediction for $X_i$ of a Gaussian process regression model fit to the $i$-th augmented LOO data set, $\augloomu{i}(X_i)$, is then given by
\begin{align}
\augloomu{i}(X_i) & = K(X_i, \mathbf{X}_{-i})[K(\mathbf{X}_{-i}, \mathbf{X}_{-i}) + \sigma^2 I]^{-1} Y^y_{-i},
\end{align}
and the model's predictive variance at $X_i$ is
\begin{align}
K(X_i, X_i) -  K(X_i, \mathbf{X}_{-i})[K(\mathbf{X}_{-i}, \mathbf{X}_{-i}) + \sigma^2 I]^{-1} K(\mathbf{X}_{-i}, X_i),
\end{align}
where the rows of the matrix $\mathbf{X}_{-i} \in \reals^{n \times p}$ are the inputs in $\augloodata{i}$, $Y^y_{-i} = (Y_{-i}, y)\in \reals^n$ is the vector of labels in $\augloodata{i}$, and $\sigma^2$ is the (unknown) variance of the label noise, whose value is set as a hyperparameter.
Note that the mean prediction is a linear function of the candidate value, $y$, which is of the same form as the ridge regression prediction in \eqref{eq:ridgepred}; furthermore, the predictive variance is constant over $y$. Therefore, we can mimic Alg. \ref{alg:ridge} to efficiently compute scores and weights by training just $n + 1$ rather than $(n + 1) \times |\mathcal{Y}|$ models.

\section{Additional details and results on designing fluorescent proteins}
\label{app:protein-results}

\paragraph{Features}

Each sequence was first represented as a length-thirteen vector of signed bits ($-1$ or $1$), each denoting which of the two wild-type parents the amino acid at a site matches.
The features for the sequence consisted of these thirteen signed bits, called the first-order terms in the main text, as well as all ${13 \choose 2}$ products between pairs of these thirteen bits, called the second-order interaction terms. 

\paragraph{Additional simulated measurement noise.}

Each time the $i$-th sequence in the combinatorially complete data set was sampled, for either training or designed data, we introduced additional simulated measurement noise using the following procedure.
Poelwijk et al. \cite{poelwijk2019} found that the Walsh-Hadamard transform of the brightness fitness landscape included up to seventh-order statistically significant terms.
Accordingly, we fit a linear model of up to seventh-order terms for each of the combinatorially complete data sets, then estimated the standard deviation of the $i$-th sequence's measurement noise, $\sigma_i$, as the residual between its label and this model's prediction.
Each time the $i$-th sequence was sampled, for either training or designed data, we also sampled zero-mean Gaussian noise with standard deviation $\sigma_i$ and added it to the $i$-th sequence's label.
This was done to simulate the fact that multiple measurements of the same sequence will yield different labels, due to measurement noise.

\begin{figure}[ht]
  \centering
  \includegraphics[width=\textwidth]{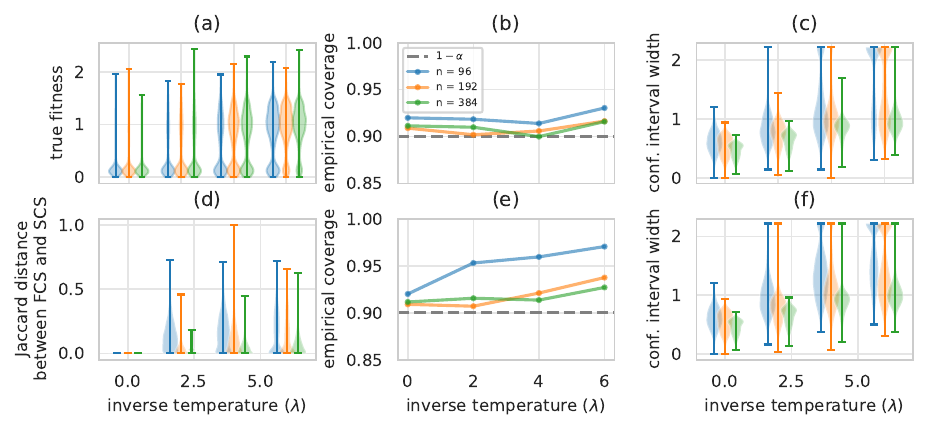}
  \caption{Quantifying predictive uncertainty for designed proteins, using the red fluorescence data set. (a) Distributions of labels of designed proteins, for different values of the inverse temperature, $\lambda$, and different amounts of training data, $n$. Labels surpass the fitness range observed in the combinatorially complete data set, $[0.025, 1.692]$, due to additional simulated measurement noise. (b) Empirical coverage, compared to the theoretical lower bound of $1 - \alpha = 0.9$ (dashed gray line), and (c) distributions of confidence interval widths achieved by full conformal prediction for feedback covariate shift (our method) over $T = 2000$ trials. (d) Distributions of Jaccard distances between the confidence intervals produced by full conformal prediction for feedback covariate shift and standard covariate shift \cite{tibshirani2019conformal}. (e, f) Same as (b, c) but using full conformal prediction for standard covariate shift. In (a), (c), (d), and (f), the whiskers signify the minimum and maximum observed values.
  }
  \label{fig:red-marginal}
\end{figure}

\begin{figure}[ht]
  \centering
  \includegraphics[width=\textwidth]{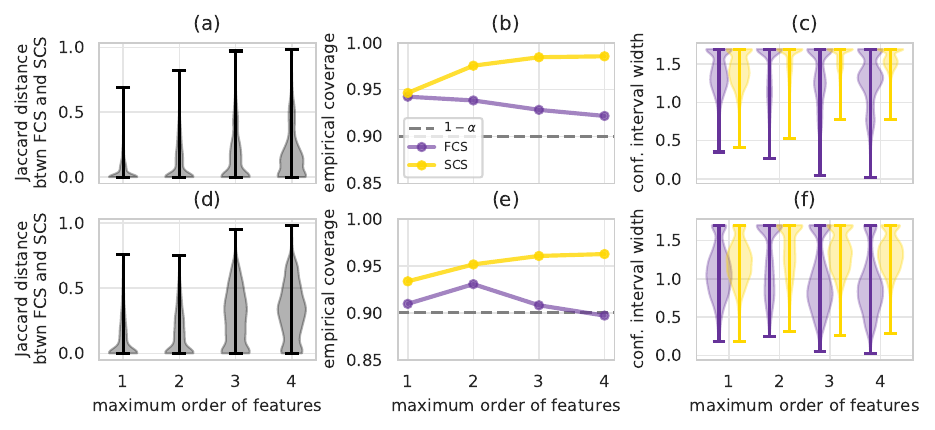}
  \caption{Quantifying predictive uncertainty for designed proteins using the blue and red fluorescence data sets, for $n = 48$ training data points, $\lambda = 6$, and ridge regression models with features of varying complexity. In particular, the features consist of all interaction terms up to order $d$ between the thirteen sequence sites, where the maximum order, $d$, is the $x$-axis of the following subplots. (a) Distributions of Jaccard distances between the confidence intervals produced by conformal prediction for feedback covariate shift (FCS, our method) and standard covariate shift (SCS) \cite{tibshirani2019conformal} for the blue data set over $T = 2000$ trials. (b) Empirical coverage, compared to the theoretical lower bound of $1 - \alpha = 0.9$ (dashed gray line), achieved by conformal prediction for FCS and SCS over those trials. (c) Distributions of confidence interval widths using conformal prediction for FCS and SCS. (d-f) Same as (a-c) but for the red fluorescence data set. In (a), (c), (d), and (f), whiskers signify the minimum and maximum observed values.}
  \label{fig:higher-orders}
\end{figure}

\begin{figure}[ht]
  \centering
  \includegraphics[width=\textwidth]{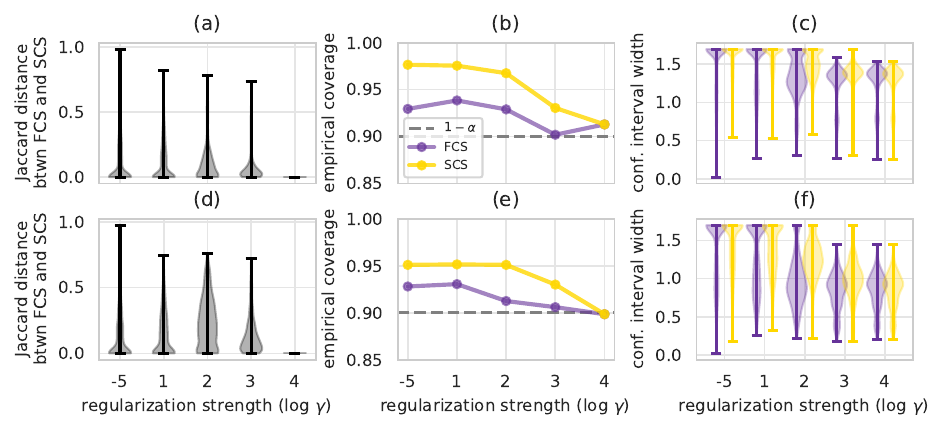}
  \caption{Quantifying predictive uncertainty for designed proteins using the blue and red fluorescence data sets, for $n = 48$ training data points, $\lambda = 6$, and varying ridge regularization strength, $\gamma$. (a) Distributions of Jaccard distances between the confidence intervals produced by conformal prediction for feedback covariate shift (FCS, our method) and standard covariate shift (SCS) \cite{tibshirani2019conformal} for the blue data set over $T = 2000$ trials. (b) Empirical coverage, compared to the theoretical lower bound of $1 - \alpha = 0.9$ (dashed gray line), achieved by conformal prediction for FCS and SCS over those trials. (c) Distributions of confidence interval widths using conformal prediction for FCS and SCS. (d-f) Same as (a-c) but for the red fluorescence data set. In (a), (c), (d), and (f), whiskers signify the minimum and maximum observed values.}
  \label{fig:regs}
\end{figure}

\begin{figure}[ht]
  \centering
  \includegraphics[width=\textwidth]{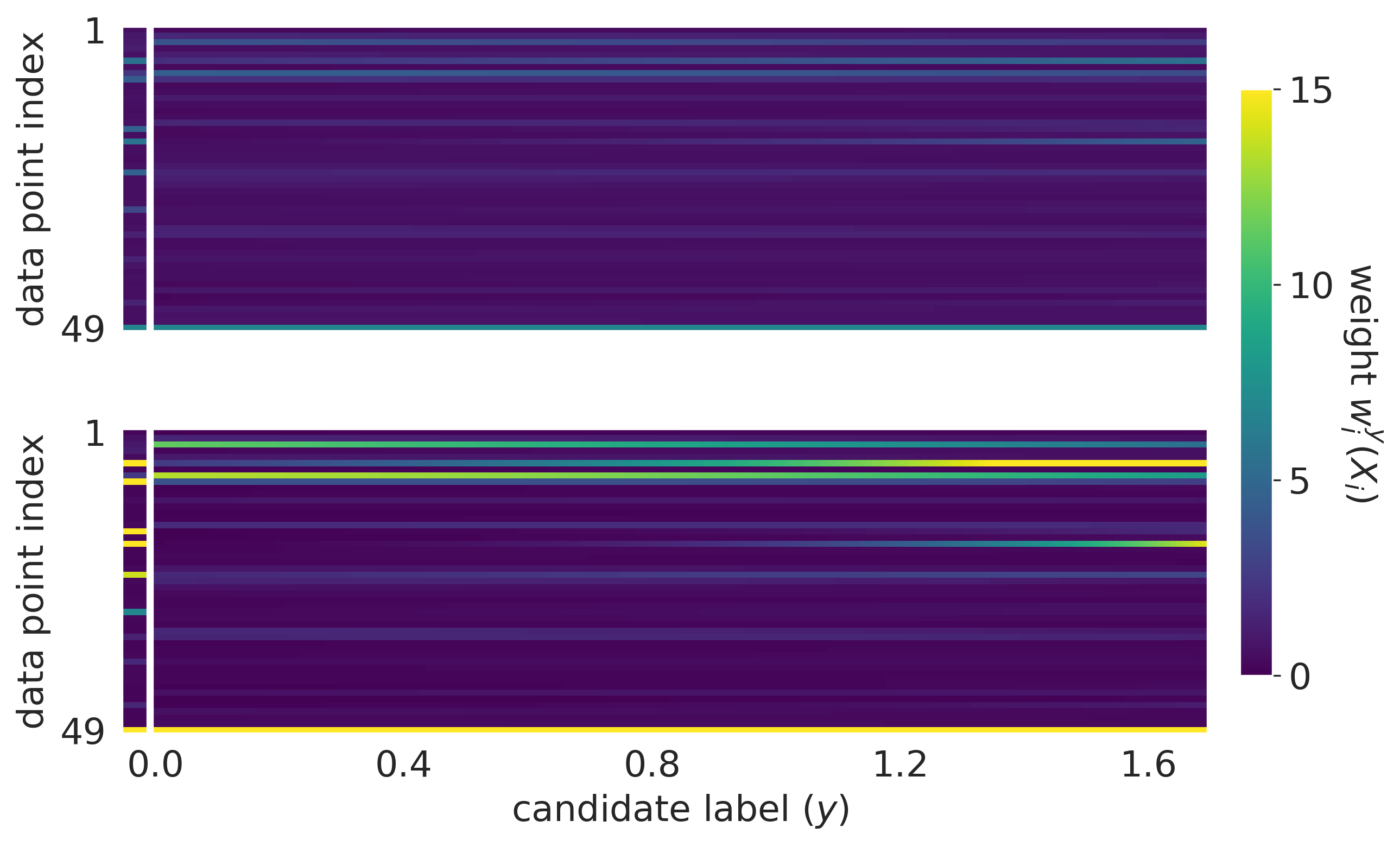}
  \caption{Comparison between the weights constructed by conformal prediction for feedback covariate shift (FSC, our method) and standard covariate shift (SCS) \cite{tibshirani2019conformal} for one example training data set and resulting designed sequence, for $n = 48$ with the blue fluorescence data set and two different settings of the inverse temperature, $\lambda$. Top: For $\lambda = 2$, vector of the $n + 1$ weights prescribed under SCS for the $n$ training data points (data point indices $1$ through $48$) and the candidate test data points (data point index $49$), alongside $(n + 1) \times |\mathcal{Y}|$ matrix of the weights prescribed under FCS for those same $n + 1$ training and candidate test data points. The weight for each of these data points depends on the candidate label, $y$ ($x$-axis of heatmap), through a linear relationship with $y$ (see Section \ref{sec:ridge}). Bottom: same as top but for $\lambda = 6$.}
  \label{fig:weights-comparison-lambda}
\end{figure}

\begin{figure}[ht]
  \centering
  \includegraphics[width=\textwidth]{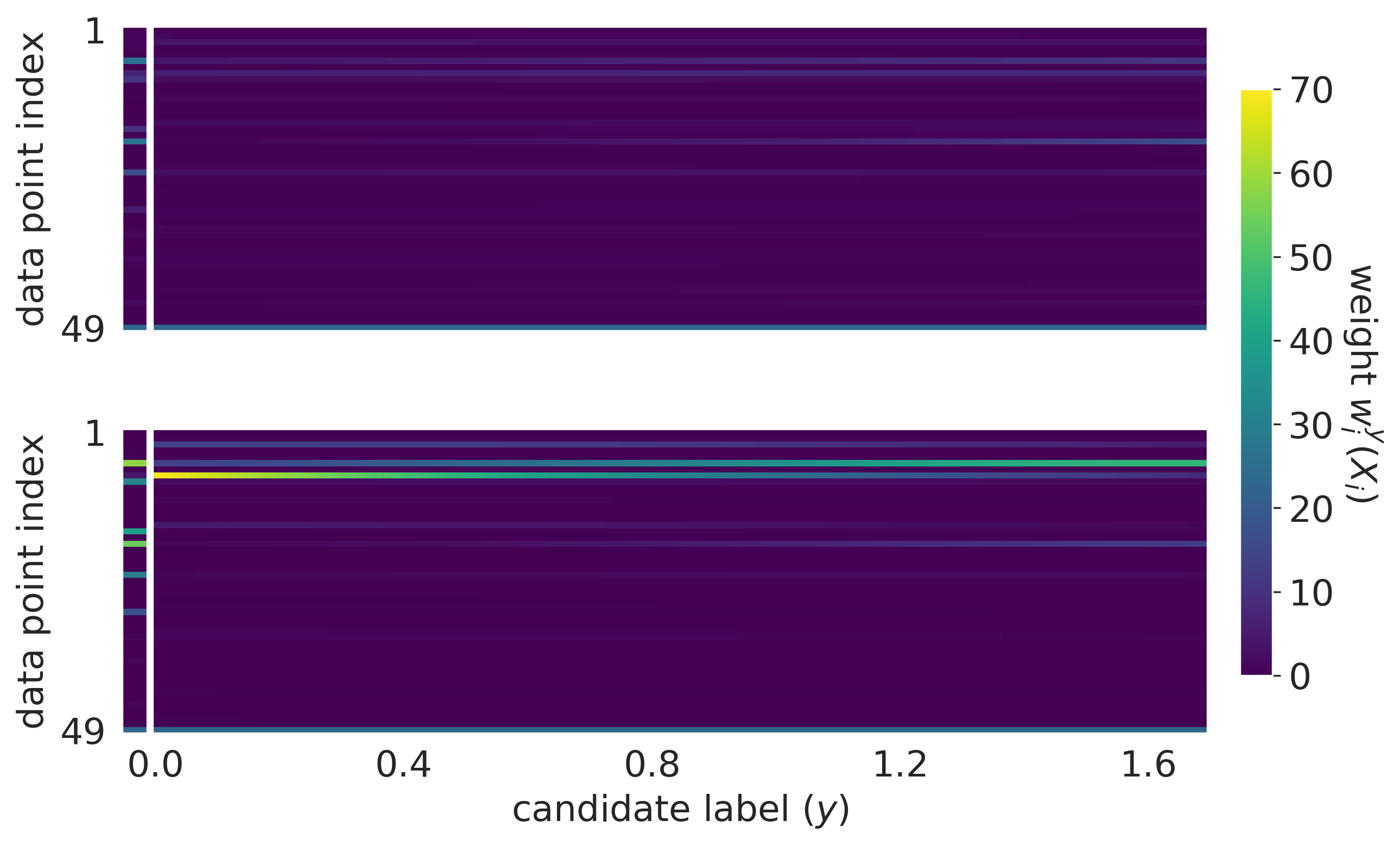}
  \caption{Comparison between the weights constructed by conformal prediction for feedback covariate shift (FSC, our method) and standard covariate shift (SCS) \cite{tibshirani2019conformal} for one example training data set and resulting designed sequence, for $n = 48$ with the blue fluorescence data set and two different settings of the ridge regularization strength, $\gamma$. Top: For $\gamma = 100$, vector of the $n + 1$ weights prescribed under SCS for the $n$ training data points (data point indices $1$ through $48$) and the candidate test data points (data point index $49$), alongside $(n + 1) \times |\mathcal{Y}|$ matrix of the weights prescribed under FCS for those same $n + 1$ training and candidate test data points. The weight for each of these data points depends on the candidate label, $y$ ($x$-axis of heatmap), through a linear relationship with $y$ (see Section \ref{sec:ridge}). Bottom: same as top but for $\gamma = 10$.}
  \label{fig:weights-comparison-reg}
\end{figure}

\section{Additional details on AAV experiments}
\label{sect:app-aav}

\paragraph{NNK sequence distribution.}
The NNK sequence distribution is parameterized by independent categorical distributions over the four nucleotides, where the probabilities of the nucleotides are intended to result in a high diversity of amino acids while avoiding stop codons.
Specifically, for three contiguous nucleotides corresponding to a codon, the first two nucleotides are sampled uniformly at random from $\{\texttt{A}, \texttt{C}, \texttt{T}, \texttt{G}\}$, while the last nucleotide is sampled uniformly at random from only $\{\texttt{T}, \texttt{G}\}$.

\paragraph{Additional simulated measurement noise.}

Following Zhu \& Brookes et al. \cite{zhu2021aav}, the fitness assigned to the $i$-th sequence was an enrichment score based on its counts before and after a selection experiment, $n_{i, \text{pre}}$ and $n_{i, \text{post}}$, respectively.
The variance of this enrichment score for the $i$-th sequence was estimated as
\begin{align}
    \sigma_i^2 = \frac{1}{n_{i, \text{post}}} \left(1 - \frac{n_{i, \text{post}}}{N_{\text{post}}} \right) + \frac{1}{n_{i, \text{pre}}} \left(1 - \frac{n_{i, \text{pre}}}{N_{\text{pre}}} \right)
\end{align}
where $N_{\text{pre}}$ and $N_{\text{post}}$ denote the total counts of all the sequences before and after the selection experiment, respectively.
Using this estimate, we introduced additional simulated measurement noise to the label of the $i$-th sequence by adding zero-mean Gaussian noise with a variance of $0.1 \cdot \sigma_i^2$.

\paragraph{Neural network details.} 

As in \cite{zhu2021aav}, the neural network took one-hot-encoded sequences as inputs and had an architecture consisting of two fully connected hidden layers, with $100$ units each and $\texttt{tanh}$ activation functions.
It was fit to the $7,552,729$ training data points with the built-in implementation of the Adam algorithm in Tensorflow, using the default hyperparameters and a batch size of $64$ for $10$ epochs, where each training data point was weighted according to its estimated variance as in \cite{zhu2021aav}.

% solving for

% rejection sampling